\documentclass[10pt,twocolumn,letterpaper]{article}
\pdfoutput=1
\usepackage{cvpr}              

\makeatletter
\@namedef{ver@everyshi.sty}{}
\makeatother
\usepackage{tikz}
\usepackage{graphicx}
\usepackage{amsmath}
\usepackage{amssymb}
\usepackage{pifont}
\newcommand{\cmark}{\ding{51}}%
\newcommand{\xmark}{\ding{55}}%
\usepackage{diagbox}
\usepackage[linesnumbered,ruled,vlined]{algorithm2e}
\usepackage{bbding}
\usepackage{booktabs}
\usepackage{utfsym}
\usepackage{makecell}
\usepackage{bm}
\usepackage[accsupp]{axessibility} 

\usepackage[pagebackref,breaklinks,colorlinks]{hyperref}

\usepackage{multirow}

\usepackage[capitalize]{cleveref}

\usepackage{amsthm}
\newtheorem{thm}{Theorem}[section]

\newtheorem{cor}[thm]{Corollary}

\crefname{section}{Sec.}{Secs.}
\Crefname{section}{Section}{Sections}
\Crefname{table}{Table}{Tables}
\crefname{table}{Tab.}{Tabs.}

\def\cvprPaperID{3847} 
\def\confName{CVPR}
\def\confYear{2023}
\begin{document}

\title{Bi-directional Distribution Alignment for Transductive Zero-Shot Learning}

\author{Zhicai Wang$^1$, Yanbin Hao$^{1*}$, Tingting Mu$^2$, Ouxiang Li$^1$, Shuo Wang$^1$, Xiangnan He$^{1}$\thanks{Yanbin Hao and Xiangnan He are both the corresponding authors.}\\
$^1$University of Science and Technology of China, $^2$The University of Manchester\\
{\tt\small wangzhic@mail.ustc.edu.cn, haoyanbin@hotmail.com, tingting.mu@manchester.ac.uk,}\\
{\tt\small  lioox@mail.ustc.edu.cn, \{shuowang.hfut,xiangnanhe\}@gmail.com}
}
\maketitle
\begin{abstract}

Zero-shot learning (ZSL) suffers intensely from the domain shift issue, i.e., the mismatch (or misalignment) between the true and learned data distributions for classes without training data (unseen classes).
By learning additionally from unlabelled data collected for the unseen classes, transductive ZSL (TZSL) could reduce the shift but only to a certain extent.
To improve TZSL, we propose a novel  approach Bi-VAEGAN which strengthens the distribution alignment between the visual space and an auxiliary space. 
As a result, it can reduce largely the domain shift. 
The proposed key designs include (1) a bi-directional distribution alignment, (2) a simple but effective $L_2$-norm based feature normalization approach, and (3) a more sophisticated unseen class prior estimation. 
%
Evaluated by four benchmark datasets,  Bi-VAEGAN\footnote{Code is available at \href{https://github.com/Zhicaiwww/Bi-VAEGAN}{https://github.com/Zhicaiwww/Bi-VAEGAN}} achieves the new state of the art under both the standard and generalized TZSL settings. 
%



\vspace{-0.2cm}
\end{abstract}
\vspace{-0.3cm}

\section{INTRODUCTION}

Zero-shot learning (ZSL) was originally known  as zero-data learning in  computer vision  \cite{larochelle2008zero}. 
The goal is to tackle the challenging training setup of largely restricted  example and (or) label availability \cite{chen2022msdn}. 
For instance, in conventional ZSL,  no training example is provided for the targeted classes, and they are therefore referred to as the unseen classes.   
What is provided instead is a large amount of  training examples paired with their class labels but for a different set of  classes  referred to as the seen classes. 
This setup is also known as the inductive ZSL. 
Its core challenge is to enable the classifier to extract knowledge from the seen classes and transfer it to the unseen classes, assuming the existence of such relevant knowledge \cite{norouzi2013zero,zhang2016zero,xian2018zero}. 
For instance,  a ZSL classifier can be constructed to recognize \textit{leopard} images after feeding it the Felinae images like \textit{wildcat}, knowing  \textit{leopard} is relevant to Felinae.  
Information on class relevance is typically provided as auxiliary data,  bridging knowledge transfer from the seen to unseen classes.
The auxiliary data can be human-annotated attribute information  \cite{wah2011caltech}, text description \cite{reed2016learning}, knowledge graph \cite{lee2018multi} or a formal description of knowledge (e.g., ontology) \cite{geng2021ontozsl}, etc., which are encoded as (set of) embedding vectors. 
Learning solely from auxiliary data to capture class relevance is challenging, resulting in a discrepancy between the true and modeled distributions for the unseen classes, known as the domain shift problem.
To ease the learning, another ZSL setup called transductive ZSL (TZSL) is proposed. 
It allows to additionally include in the training unlabelled examples collected for the targeted classes.
Since it does not require any extra annotation effort to pair the examples from the unseen classes with their class labels, this setup is still practical in real-world applications.

\begin{figure}[t]
\begin{center}
    \includegraphics[width = 1\linewidth,height = 5.3cm]{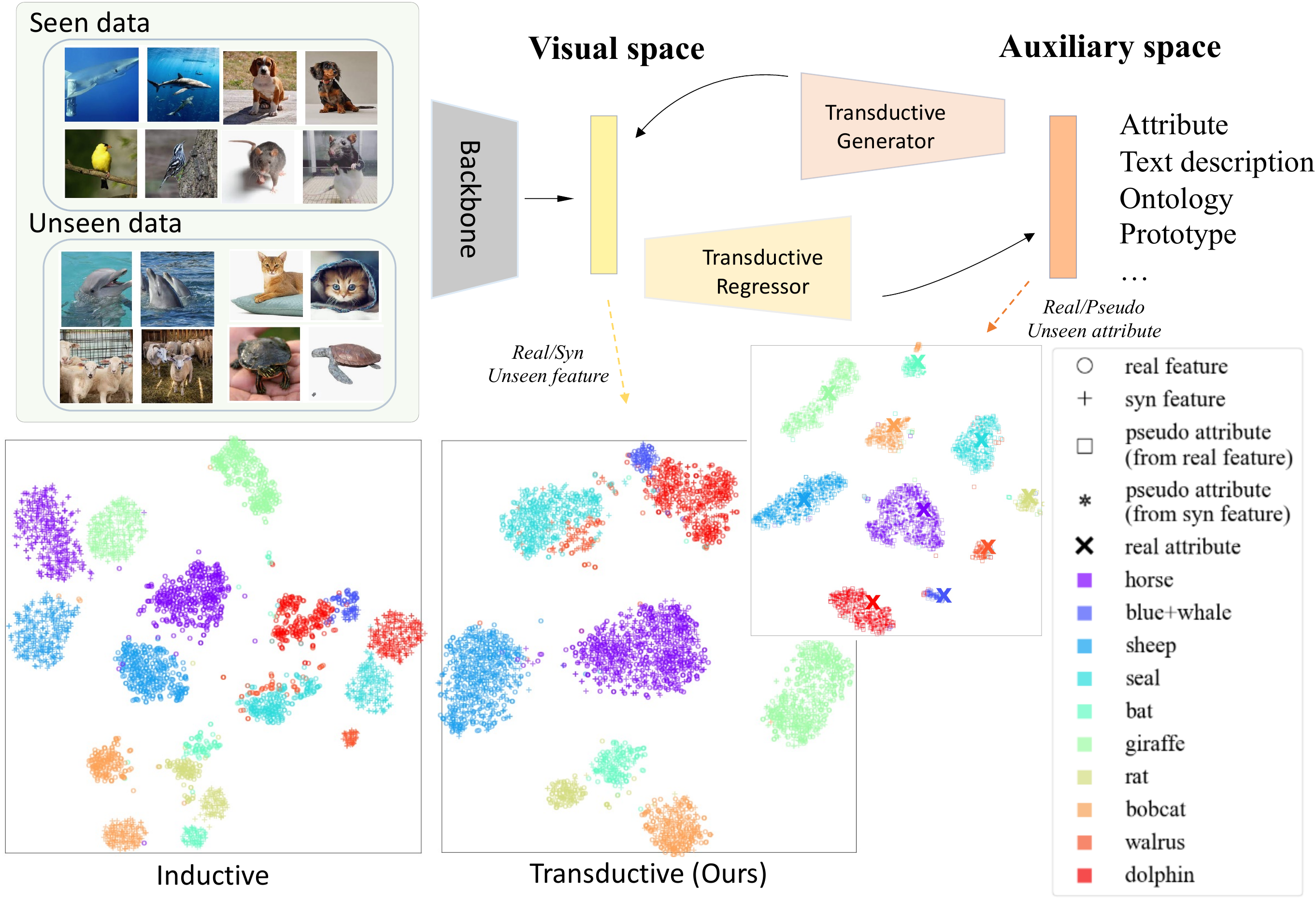}
\end{center}    
\caption[]{The top figure illustrates the proposed bi-directional generation between the visual and auxiliary spaces. The bottom figure compares the aligned visual space obtained by our method with the unaligned one obtained by inductive ZSL for the unseen classes using the AWA2 data. The bottom right figure shows our approximately aligned attributes in auxiliary space.   \label{fig.1}}
\vspace*{-0.4cm}
\end{figure}

Supported by sufficient data examples, generative models have become popular, for instance, used to synthesize examples to enhance classifier training \cite{xian2018feature,elhoseiny2019creativity,li2019leveraging} and to learn the unseen data distribution \cite{wu2020self,paul2019semantically,xian2019f} under the TZSL setup. 
Depending on the label availability, they can be formulated as unconditional generation i.e., $p(\bm v)$, or conditional generation i.e., $p(\bm v|\bm y)$.
When being conditioned on the auxiliary information, which is a more informative form of class labels, a data-auxiliary (data-label) joint distribution can be learned.
Such distributions  bridge the knowledge between the visual and auxiliary spaces and enables the generator to be a powerful tool for knowledge transfer.
By including  an appropriate supervision,  e.g, by using a conditional discriminator to discriminate whether the generation is a realistic \textit{wildcat} image, intra-class data distributions can be aligned with the real ones.
However, the challenge of TZSL is  to transfer the knowledge contained by the joint data-auxiliary distribution for the seen classes to improve the distribution modelling  for the unseen classes, and achieve a realistic generation for the unseen classes.
A representative generative approach for achieving this is f-VAEGAN \cite{xian2019f}. 
It enhances the unseen generation using an unconditional discriminator and learns the overall unseen data distribution.
This simple strategy turns out to be effective in approximating the conditional distribution of unseen classes.
Most existing works use auxiliary data in the forward generation process  \cite{wu2020self,bo2021hardness}, i.e.,  to generate images from the auxiliary data as by $p(\bm v|\bm y)$. 
This can result in weakly guided conditional generation for the unseen classes and the alignment is extremely sensitive to the quality of the auxiliary information.
To bridged better between the visual and auxiliary data, particularly for the unseen classes, which is equivalent to enhancing the alignment with the true unseen conditional distribution, we propose a novel bi-directional generation process. 
It couples the forward generation process with a backward one, i.e., to generate auxiliary data from the images  as by $p(\bm y|\bm x)$.

Figure \ref{fig.1} illustrates our proposed bi-directional generation based on a feature-generating framework. It builds upon the baseline design  f-VAEGAN, and is named as Bi-VAEGAN.
Overall, the proposed design covers three aspects.  (1) A transductive regressor is added to form the backward generation, synthesizing pseudo auxiliary features conditioned on visual features of an image. 
This, together with the forward generation as used in  f-VAEGAN, provides more constraints to learn the unseen conditional distribution, expecting to achieve better alignments between the visual and auxiliary spaces.
(2) We introduce  $L_2$-feature normalization, a free-lunch data pre-processing method,  to further support the conditional alignment. 
(3) Besides, we note that the (unseen) class prior plays a crucial role in distribution alignment, particularly for those datasets that have extremely unbalanced label distribution. 
A poor choice of the class prior can easily lead to a poor alignment. 
To address this issue, we propose a simple but effective class prior estimation approach based on a cluster structure contained by the examples from the unseen classes.  
The proposed Bi-VAEGAN is compared against various advanced ZSL techniques using four benchmark datasets and achieves a significant TZSL classification accuracy boost compared with the other generative benchmark models.
\section{RELATED WORKS}
\noindent{\textbf{Inductive Zero-Shot Learning }}
Previous works on inductive ZSL learn simple projections from the auxiliary (e.g. semantic)  to visual spaces to enable knowledge transfer from the seen to unseen classes \cite{changpinyo2017predicting,zhang2017learning,DBLP:conf/nips/YuJFGPZ18,chou2020adaptive}.
They suffer from the domain shift problem due to the distribution gap between the seen and unseen data. 
Relation-Net \cite{sung2018learning} utilizes two embedding modules to align the visual and semantic information, modelling relations  in the embedded space.
Generative approaches, e.g., variational auto-encoder (VAE) \cite{doersch2016tutorial} and generative adversarial network (GAN) \cite{creswell2018generative},  synthesize unseen examples and train an additional classifier using the generated examples. Although this improves alignment between the synthesized and true  distributions for the unseen classes, it still suffers from domain shift. 
Some works, on the other hand, improve by introducing auxiliary modules. For instance, f-CLSWGAN \cite{xian2018feature}  uses a Wasserstein GAN (WGAN) to model the conditional distribution of the seen data, and introduces a classification loss to improve the generation. 
Cycle-WGAN \cite{felix2018multi} employs a semantic regressor with a cycle-consistency loss \cite{zhu2017unpaired} to reconstruct the original semantic features, which provides stronger generation constraints and shares some similar spirit to our work. 
However, because there is no knowledge of the unseen data, the inductive ZSL heavily relies on the quality of the auxiliary information, making it challenging to overcome performance bottleneck.

\noindent{\textbf{Transductive Zero-Shot Learning }}
As a concession of inductive ZSL, TZSL uses test-time unseen data to improve  training \cite{fu2015transductive,wu2020self,yang2022iterative}. 
A representative approach is visual structure constraint (VSC) \cite{wan2019transductive}. 
It exploits the cluster structure of the unseen data and proposes to align the projection centers with the cluster centers. 
Recently, generative models have been actively explored and shown superiority in TZSL. 
For instance, f-VAEGAN combines  VAE and GAN, and includes an additional unconditional discriminator to capture the unseen distribution. 
SDGN \cite{wu2020self} introduces a self-supervised objective to mine discriminability  between the seen and unseen domains. 
STHS-WGAN \cite{bo2021hardness} iteratively adds easily distinguishable unseen classes to the training  examples of the seen classes to improve the unseen generation. 
However, these previous approaches mostly work with a uni-directional generation from the auxiliary to visual spaces. 
This could potentially result in limited constraints when learning unseen distributions. 
TF-VAEGAN \cite{narayan2020latent} enhances the generated visual features by utilizing an inductive regressor trained with seen data and a feedback module. Expanding this idea, our work explore additionally the unseen data information in the regressor.

\subsection{Notations}
We use $V^s=\{\bm v^s_i\}_{i = 1}^{n_s}$ and $V^u=\{\bm v^u_i\}_{i = 1}^{n_u}$ to denote the collections of examples from the seen and unseen classes, where each example is characterized by its visual features extracted by a pre-trained network.  
For examples from the seen classes, their class labels are provided and denoted by  $Y^s = \{y_i \}_{i = 1}^{n_s}$. 
Attributes (we set as the default choice of auxiliary information) are provided to describe both the seen and unseen classes, represented by vector sets $A^s=\{\bm a^s_i\}_{i = 1}^{N_s}$ and $A^u=\{\bm a^u_i\}_{i = 1}^{N_u}$ where $N_s$ and $N_u$ are the numbers of the seen and unseen classes. 
Under the TZSL setting, a classifier $f(\bm v): \mathcal{V}^u \rightarrow \mathcal{Y}^u$ is trained to conduct inference on unseen data, where we use $\mathcal{V}$ to denote the  visual representation (feature) space, and  $\mathcal{Y}$ the label space.
The  training pipeline learns from information provided by $D^{tr}=\{\langle V^s,Y^s\rangle,V^u,\{A^s, A^u\}\}$, where we use $\langle \cdot, \cdot \rangle$ to highlight the 
paired data.

\subsection{$L_2$-Feature Normalization}

Feature normalization is an important data preprocessing method, which can improve the model training and convergence\cite{li2021feature}. 
A common practice in TZSL is to normalize the visual features by the Min-Max approach i.e.,$\bm v^\prime = \frac{\bm v -\text{min}(\bm v)}{\text{max}(\bm v)-\text{min}(\bm v)}$\cite{narayan2020latent,xian2019f,wu2020self}. 
However, we find that it suffers from the distribution skew when processing the synthesized features and it is more beneficial to normalize the visual features by their $L_2$-norm. 
For a visual feature vector $\bm v  \in {V}^s \bigcup {V}^u $, it has
\vspace{-0.3cm}
\begin{equation}
\label{V_norm}
\bm v ^\prime = L_2(\bm v, r) = \frac{r\bm v}{||\bm v||_2},
\end{equation}
where the hyperparameter $r>0$ controls the norm of the normalized feature vector.  
As a result, we replace in the generator  its last $sigmoid$  layer  that accompanys the Min-Max approach with an $L_2$-normalization layer. 
We discuss further its effect in Section \ref{sec:norm}.

\begin{figure*}[thp]
    \small
    \centering
        \includegraphics[width=0.9\linewidth,height = 7.6cm]{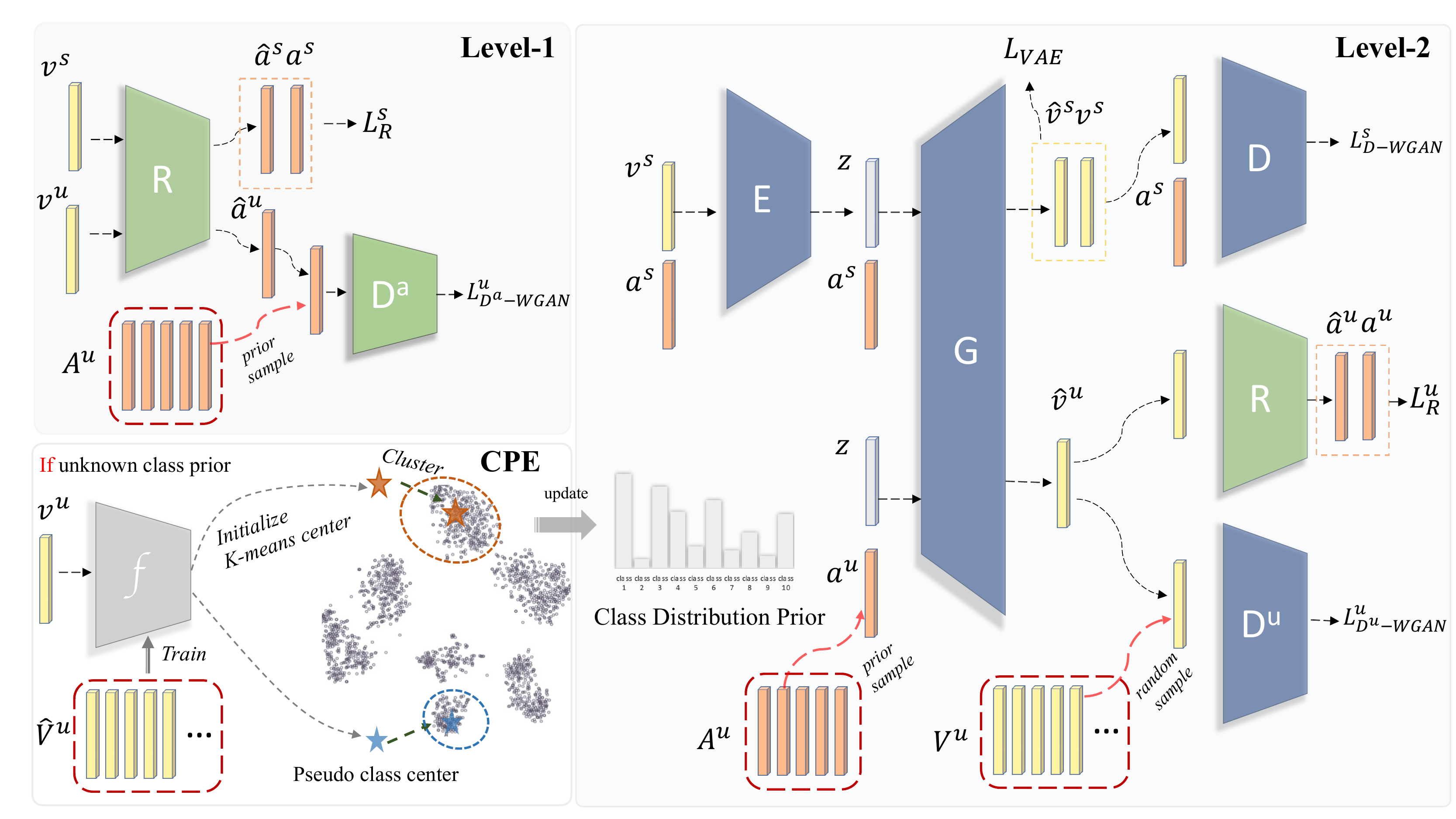} 
        \caption{The proposed Bi-VAEGAN model architecture under the TZSL setup.}
        \vspace{-0.5cm}
        \label{fig.3}
\end{figure*}

\subsection{Bi-directional Alignment Model}
\label{sec:BA}

We propose a modular model architecture composed of six modules: (1) a conditional VAE encoder $\bm E: \mathcal{V} \times 
\mathcal{A} \rightarrow \mathbb{R}^k$ mapping the visual features  to a $k$-dimensional hidden representation vector conditioned on the class attributes,  (2) a  conditional visual generator $\bm G: \mathcal{A} \times \mathbb{R}^k  \rightarrow \mathcal{V}$ synthesizing visual features from  a $k$-dimensional random  vector that is usually sampled from a normal distribution $\mathcal{N}(\bm 0, \bm 1)$, conditioned on the class attributes, 
(3) a conditional  visual Wasserstein GAN (WGAN)  critic $D: \mathcal{V}^s\times \mathcal{A } \rightarrow \mathbb{R}$ for the seen classes,  (4) a visual WGAN critic $ D^{u}: \mathcal{V}^u\rightarrow \mathbb{R}$ for the unseen classes, (5) a regressor mapping the visual space to the attribute space  $\bm R: \mathcal{V}\rightarrow \mathcal{A }$, and (6) an attribute WGAN critic $D^{a}: \mathcal{A}\rightarrow \mathbb{R}$. 
%

The proposed workflow consists of two levels. In \textbf{Level-1}, the regressor $\bm R$ is adversarially trained using the critic $D^{a}$ so that the pseudo attributes converted from the visual features align with the true attributes. 
In \textbf{Level-2}, the visual generator $\bm G$ is adversarially trained using the two critics $D$ and $D^{u}$ so that the generated visual features align with the true visual features. 
Additionally, the training of $\bm G$ depends on the regressor $\bm R$. 
This encourages the pseudo attributes converted from the synthesized visual features align better with the true attributes.
To highlight the core innovation of aligning true and fake data in both the visual and attribute spaces, we name the proposal as bi-directional alignment.

\subsubsection{Level 1: Regressor Training}
\label{sec:3.3.1}
The regressor $\bm R$ is trained transductively and adversarially.  It is constructed by performing supervised learning using the labeled examples from the seen classes, additionally enhanced by unsupervised learning from the visual features and class attributes of the unseen classes. By  ``unsupervised", we mean that the features and classes are unpaired for examples from the unseen classes.
For each example from the seen classes, $\bm R$ learns to map its visual features close to its corresponding class attributes via minimizing
\vspace{-0.3cm}
\begin{align}
    \begin{split}
        \small
         L_R^s({\mathcal{A}}^s, \mathcal{V}^s ) =&  \mathbb{E}[\| \bm R (\bm v^s) - {\bm a}^s \|_1].\\  
    \end{split}
\end{align}
Simultaneously, for examples from the unseen classes, it learns to distinguish their true attributes from the pseudo ones computed from the real unseen visual features by maximizing the adversary objective
\begin{align}
\label{adv_obj}
    \begin{split}
    \small
         L^{u}_{D^a\text{-WGAN}} ({\mathcal{A}}^u, \mathcal{V}^u ) =&  \mathbb{E} \left [D^{a}({\bm a}^u) \right ]  - \mathbb{E} \left [D^{a}(\hat{\bm a}^u) \right ]+\\  
         & \mathbb{E}_{}  [(\|\nabla_{\bar{\bm a}^u} D^{a}(\bar{\bm a}^u)\|_2-1)^2 ],
    \end{split}
\end{align}
where $\hat{\bm a}^u =\bm R (\bm v^u)$,  $\bm a^u \sim p^u_{\bm G}( y)$ and $\bar{\bm a}^u\sim \mathcal {P}_{t}({\bm a}^u,\hat{\bm a}^u)$\footnote{$\mathcal {P}_{t}(\bm a,\bm b)$ is an interpolated distribution in the $L_2$ hypersphere. An example sampled from this distribution is computed by $c = L_2(t\bm a+(1-t)\bm b,r)$ with $t\sim \mathcal{U}(0,1)$ where $\|\bm a\|_2=\|\bm b\|_2=r$.}. 
Note that the original  attribute vectors are sampled from  the unseen class prior $p^u_{\bm G}( y)$ explained in Section \ref{sec:prior}, and we refer to this as the \textbf{prior sample} process.
The third term in Eq. (\ref{adv_obj}) is known as the gradient penalty \cite{gulrajani2017improved},
which enables the Lipschitz restriction in the original WGAN \cite{arjovsky2017wasserstein}.

The regressor $\bm R$ aims at learning a mapping from the visual to the attribute features for the seen classes in a supervised manner, and meanwhile learning the distribution of the overall feature domain for the unseen classes in an unsupervised manner. The  level-1 training is formulated by
\begin{align}
\label{level1_training}
    \begin{split}
        \operatorname*{min}_{\bm R}\operatorname*{max}_{D^a}\; L_{R}^{s}+\lambda L_{D^a\text{-WGAN}}^{u},
    \end{split}
\end{align}
where $\lambda$ is a hyper-parameter. It enables knowledge transfer from the seen to  unseen classes in the attribute space, where the feature discriminability  is however limited by the hubness problem \cite{zhang2017learning}. 
But it serves as a good auxiliary module to provide ``approximate supervision'' for the unseen distribution alignment later in the visual space.

\subsubsection{Level 2: Generator and Encoder Training}  

The generator $\bm G$ is also trained transductively and adversarially. It aims at aligning the synthesized and true features,   using the visual critics $D$ and $D^u$ in the visual space, while  using the frozen regressor $\bm R$ in the attribute space.

The two visual critics are trained to get better at distinguishing the true visual features from the synthesized ones computed by the conditional generator, i.e. $ \hat{\bm v} \sim  \bm G(\bm a,\bm z) $ where $\bm z \sim \mathcal{N}(\bm 0,\bm 1)$ and $\bm a \sim p_{\bm G}(y)$. 
For the seen classes, their class prior, denoted by $p_{\bm G}^{s}(y)$\footnote{For the intra-class alignment, i.e, we know the paired seen labels, the choice of $p_{\bm G}^{s}(y)$ will not affect much of the training.}, is simply estimated from the number of examples collected for each class. 
For the unseen classes, the estimated class prior $p_{\bm G}^u(y)$ is used.
The synthesized visual feature $\hat{\bm v}$ is already $L_2$-normalized. 
For the seen classes, the critic is conditioned on their class attributes, i.e., $D(\hat{\bm v}^s, {\bm a}^s)$, while  for the unseen classes, the critic is unconditional, i.e., $D^u(\hat{\bm v}^u)$. 
The two critics $D$ and $D^u$ are trained based on the adversarial objectives:
\begin{align}
\label{Du_training1}
    \begin{split}
    \small
     L^s_{D\text{-WGAN}}(A^s, V^s) =& \mathbb{E}[D( {\bm v}^s,  {\bm a}^s)]- \mathbb{E}[D(\hat{\bm v}^s, {\bm a}^s)] +\\ 
     & \mathbb{E}[(\|\nabla_{\bar{\bm v}^s} D(\bar{\bm v}^s, {\bm a}^s)\|_2-1)^2],
    \end{split}
\end{align}
and 
\begin{align}
    \label{Du_training}
        \begin{split}
        \small
            L^{u}_{D^u\text{-WGAN}}(A^u, V^u)  = &\mathbb{E}[D^{u}({\bm v}^u )]- \mathbb{E}[D^{u}( \hat{\bm v}^u )]+ \\ 
         & \mathbb{E}[(\|\nabla_{\bar{\bm v}^u} D^u(\bar{\bm v}^u)\|_2-1)^2],
        \end{split}
    \end{align}
where $\bar{\bm v}^s$ and $\bar{\bm v}^u$ are sampled from the interpolated distribution as explained in footnote 1. Here, $\hat{\bm v}^u$ is computed from the unseen attributes sampled by $\bm a^{u} \sim p_{\bm G}^{u}(y)$.
The critic  $D^{u}$ in Eq. (\ref{Du_training}) captures the Earth-Mover distance over the unseen data distribution.

Eqs (\ref{Du_training1}) and (\ref{Du_training}) weakly align the conditional distribution of the unseen classes. It suffers from the absence of any supervised constraints. 
To further  strengthen the alignment, we introduce another training loss, as
\begin{align}
\label{vae_loss}
    \begin{split}
         L_R^u(A^u ) =&  \mathbb{E}[\| \bm R (\bm G({\bm a}^u,\bm z)) - {\bm a}^u \|_1].\\  
    \end{split}
\end{align}
It employs the regressor $\bm R$ trained in level-1 to  enforce supervised constraints.
As shown in f-VAEGAN, feature-VAE has the potential of preventing model collapse, and it could serve as a suitable complement to GAN training. Similarly, we adopt the VAE objective function to enhance the adversarial training over the seen classes:
\begin{align}
    \begin{split}
    \small
        L^s_{\text{VAE}}(A^s,& V^s)
    =   \mathbb{E} [\text{KL}( \bm E({\bm v}^s,{\bm a}^s)  \|\mathcal{N}(\bm 0, \bm 1))] + \\
        &\mathbb{E}_{\bm z^s \sim \bm E({\bm v}^s,{\bm a}^s)} [ \left(\|\bm G({\bm a}^s,\bm z^s)-{\bm v}^s\|^2_2 \right)].  \label{loss:VAE}
    \end{split}
\end{align}
The first term is the Kullback-Leibler divergence and the second is the mean-squared-error (MSE) reconstruction loss using the $L_2$-normalized feature. 
Finally,  the  level-2 training is formulated by,
\begin{align}
\label{level2_training}
        \operatorname*{min}_{\bm E,\bm G}\operatorname*{max}_{D,D^u}\; L^s_{\text{VAE}} + \alpha L_{D\text{-WGAN}}^{s} + \beta L_{R}^{u}+ \gamma L_{D^u\text{-WGAN}}^{u},
\end{align}
where $\alpha$, $\beta$ and $\gamma$ are hyper-parameters. It  transfers the knowledge of \textit{ paired visual features and attributes} of the seen classes and the estimated \textit{class prior} of the unseen classes, 
and   is enhanced by  the attribute regressor $\bm R$ to constrain further the visual feature generation for unseen classes. 
The proposed Bi-VAEGAN architecture can be easily modified to accommodate the inductive ZSL setup, by removing all the loss functions using the unseen data $V^u$. This results in the following:
\begin{align}
\label{inductive_training}
    \begin{split}
        \textmd{For level-1: } &\min_{\bm R} L_{R}^{s} \\
         \textmd{For level-2: }&\min_{\bm E,\bm G}\operatorname*{max}_{D} L^s_{\text{VAE}} + \alpha L_{D\text{-WGAN}}^{s} + \beta L_{R}^{u}.
    \end{split}
\end{align}

\subsubsection{Unseen Class Prior Estimation}
\label{sec:prior}

When training based on the objective functions in Eqs. (\ref{adv_obj}) and (\ref{Du_training}), the attributes for the unseen classes are sampled from the class prior:  $\bm a^u \sim p^u_{\bm{G}}( y)$. Since there is no label information provided for the unseen classes, it is not possible to sample from the real class prior $p^u(y)$. An alternative way to estimate $p^u_{\bm{G}}( y)$ is needed.
We have observed that examples from the unseen classes possess fairly separable cluster structures in the visual space thanks to the strong backbone network. Therefore, we propose to estimate the unseen class prior based on such cluster structure.
We employ the K-means clustering algorithm and carefully design the initialization of its cluster centers since the prior estimation is   sensitive to the initialization.
The estimated prior $p_{\bm G}^u(y)$ is iteratively updated and in each epoch cluster centers are re-initialized by the pseudo class centers calculated from an extra classifier $f$. 
For the very first estimation of $p_{\bm G}^u(y)$, rather than using the naive but sometimes (if it differs greatly from the real prior) harmful uniform class prior,  we use the inductively trained generator to transfer the seen paired knowledge to have a better informed estimation for the unseen classes.
We refer to this estimation approach as the cluster prior estimation (CPE), and its implementation is shown in Algorithm \ref{tioncpe} (lines 1-12).

\IncMargin{1em}
\begin{algorithm}[tp]\SetAlgoLined
     \SetKwData{Left}{left}\SetKwData{This}{this}\SetKwData{Up}{up} \SetKwFunction{Union}{Union}\SetKwFunction{FindCompress}{FindCompress} \SetKwInOut{Data}{Data}\SetKwInOut{Input}{Input}\SetKwInOut{Output}{Output}
    \small
    \Input{$\langle V^s,Y^s \rangle$,  $V^u$,  $\{A^s, A^u\}$, unseen class number $N_u$, epoch numbers $T_1$ and $T_2$.}
    \Output{$\bm E$, $\bm G$, $\bm R$, $D$, $D^u$, $D^a$.}
    \BlankLine
    \For{ $i = 1$  to $T_1$}{
         Inductive training with $\langle V^s,Y^s \rangle,\{A^s, A^u\}$ by  \\Eq. (\ref{inductive_training})\;
}
      
    \For{ $i = 1$  to $T_2$}{
        Define uniform distribution label set  $Y^u_{\bm G}$\;
        Synthesize paired unseen set  $\langle \hat V^u_{\bm G}, Y^u_{\bm G}\rangle$ using $\bm G$\;
        Train a classifier $f$ using $\langle \hat V^u_{\bm G}, Y^u_{\bm G}\rangle$\; 
        Assign pseudo class labels by $\hat Y^u = f(V^u)$\;
        Compute pseudo class centers $C^u \leftarrow  \langle V^u, \hat Y^u\rangle$\;
     
      \emph{$\hat Y_{kmeans}^u$ = Kmeans($V^u$,$N_u$, InitCenter = $C^u$)}\; 
      $p_{\bm G}^u(y) \leftarrow  \hat Y_{kmeans}^u$\tcc*{\footnotesize{Update prior}}
      Transductive training with $\langle V^s,Y^s \rangle$, $V^u$, $\{A^s, A^u\}$ and $p_{\bm G}^u(y)$ by Eqs (\ref{level1_training}) and (\ref{level2_training})\;
      }
    \caption{Bi-VAEGAN (CPE) }
           \label{tioncpe} 
      \end{algorithm}
 \DecMargin{1em}

\noindent \textbf{Discussion:} We attempt to explain the importance of estimating $p^u_{\bm{G}}( y)$ based on the following corollary, which is a natural result of Theorem 3.4 of \cite{tachet2020domain}.
\begin{cor}
Under the generative TZSL setup, for the unseen classes, the total variation distance between the true conditional visual feature distribution $p^u( \bm{v}|y)$ and the estimated one by the generator $p_{\bm{G}}^u(\hat{\bm{v}}|y)$ is upper bounded by
    \begin{align}
        \begin{split}
            &\operatorname*{max}_{y \in Y^u}d_{\mathrm{TV}}\left(p^u_{\bm G}\left(\hat{\bm v}\mid y\right),p^u\left(\bm v\mid y\right) \right)\\
            \leq& \frac{1}{\min_{y \in Y^u} p^u(y)}\left( \max_{y\in Y^u}\left(\frac{p^u(y)}{p^u_{\bm{G}}(y)}\right) e^u_f(\hat{\bm v})+e^u_f(\bm v)\right. \\
            & \left.+\sqrt{8{d}_{\textmd{JS}}\left( \sum_{ y \in Y^u } p^u(y)p^u_{\bm G}(\hat{\bm v}|y), p^u(\bm v)\right)} \right),
    \end{split}
    \label{gls}
    \end{align}
    where ${d}_{\textmd{JS}}(\cdot, \cdot)$ is the Jensen-Shanon divergence between two distributions, and $e^u_f(\bm x)$ denotes the error probability that the classification  of the input feature vector disagrees with its ground truth using hypothesis $f$.

    \end{cor}

\noindent    
The above result is a straightforward application of the domain adaptation result in Theorem 3.4 of \cite{tachet2020domain}, obtained by treating $p^u_{\bm G}\left(\hat{\bm v}\mid y\right)$ as the source domain distribution while $p^u\left(\bm v\mid y\right)$ as the target domain distribution. When the estimated  and ground truth class priors are equal, i.e., $p^u(y) = p^u_{\bm{G}}(y)$, Eq. (\ref{gls}) reduces to 
\vspace*{-0.08cm}
  \begin{align}
        \begin{split}
            &\operatorname*{max}_{y \in Y^u}d_{\mathrm{TV}}\left(p^u_{\bm G}\left(\hat{\bm v}\mid y\right),p^u\left(\bm v\mid y\right) \right)\\
            \leq& \frac{1}{\gamma}\left( e^u_f(\hat{\bm v})+e^u_f(\bm v) 
            +\sqrt{8{d}_{\textmd{JS}}\left(   p^u_{\bm G}(\hat{\bm v}), p^u(\bm v)\right)} \right),
    \end{split}
    \end{align}
where $\gamma=\min_{y \in Y^u} p^u(y)$. The effect of the class information is completely removed from the bound. As a result, the success of the conditional distribution alignment is dominated by matching the unconditional distribution in $D^u$. This is important  for our model because of the unsupervised learning nature for the unseen classes. 
\vspace*{-0.2cm}

\subsubsection{Predictive Model and Feature Augmentation}
\label{sec:3.3.4}

After completing the training of the six modules, a predictive model for classifying the unseen examples is trained. This results in a classifier $f: \mathcal{V}^u (\textmd{ or } \mathcal{\hat{V}}^u) \times \mathcal{H}^u \times \hat{\mathcal{A}}^u \rightarrow \mathcal{Y}^u$ working in an augmented multi-modal  feature space \cite{narayan2020latent}. Specifically, the used feature vector $\bm x^u $  concatenates the visual features $\bm v^u$ (or $\hat{\bm v}^u $),  the pseudo attribute vector computed by the  regressor $\hat{\bm a}^u=\bm R\left(\bm v^u\right)$ and the hidden representation vector $\bm h^u$ returned by the first fully-connected layer of regressor, which gives  $\bm x^u =\left[\bm v^u, \bm h^u,  \hat{\bm a}^u \right]$.  
It integrates knowledge of the generator and regressor that is  transductive, and presents stronger discriminability.

\begin{table*}[thp]
  \centering
  \footnotesize
   \renewcommand{\multirowsetup}{\centering}

   \resizebox{\textwidth}{!}{\begin{tabular}{l|l|cccc|ccc|ccc|ccc|ccc}
      \hline
      \multicolumn{2}{c|}{\multirow{3}{*}{\textbf{Method}}}& \multicolumn{4}{c}{Zero-Shot Learning}& \multicolumn{12}{|c}{Generalized Zero-Shot Learning}\\\cline{3-18}
      \multicolumn{2}{l|}{}&AWA1&AWA2&CUB&SUN&\multicolumn{3}{c}{AWA1}&\multicolumn{3}{|c}{AWA2}&\multicolumn{3}{|c|}{CUB}&\multicolumn{3}{c}{SUN}\\\cline{3-18}
      \multicolumn{2}{l|}{}&T1&T1&T1&T1&S&U&H&S&U&H&S&U&H&S&U&H\\\hline
      \multirow{5}{*}{I}&F-CLSWGAN\cite{xian2018feature}&59.9&62.5&58.1 &54.9& 76.1&16.8&27.5&81.8&14.0&23.9&33.1&21.8&26.3&63.8&23.7&34.4\\
      &SP-AEN\cite{chen2018zero}&-&58.5&59.2&55.4&-&-&-&\underline{90.9}&23.3&37.1&38.6&24.9&30.3&\textbf{70.6}&34.7&46.6\\
      &DEM\cite{zhang2017learning}&68.4& 67.2& 61.9 &51.7& 32.8&84.7&47.3&86.4&30.5&45.1&25.6&34.3&20.5&54.0&19.6&13.6\\
      &ALE\cite{akata2013label}&68.2& -& 60.8 &57.3& 61.4&57.9&59.6&68.9&52.1&59.4&36.6&42.6&39.4&57.7&43.7&49.7\\
      &LisGAN\cite{li2019leveraging}&70.6& -& 61.7 &58.8& 76.3&52.6&62.3&-&-&-&37.8&42.9&40.2&57.9&46.5&51.6 \\\hline
      \multirow{13}{*}{T}&GMN \cite{sariyildiz2019gradient}&82.5&-&64.6&64.3&79.2&70.8&74.8&-&-&-&70.6&60.2&65.0&40.7&57.1&47.5\\
      &DSRL\cite{ye2017zero}&74.7&72.8&56.8&48.7&74.7&20.8&32.6&-&-&-&25.0&17.7&20.7&39.0&17.3&24.0\\
      &GFZSL\cite{verma2017simple}&48.1&78.6&50.0&64.0&67.2&31.7&43.1&-&-&-&45.8&24.9&32.2&-&-&-\\
      &ALE\_trans\cite{akata2013label}&-&70.7&54.5&55.7&-&-&-&73.0&12.6&21.5&45.1&23.5&30.9&22.6&19.9&21.2\\
      &PREN\cite{ye2019progressive}&-&78.6&66.4&62.8&-&-&-&88.6&32.4&47.4&55.8&35.2&43.1&27.2&35.4&30.8\\
      &f-VAEGAN$^\dagger$ \cite{xian2019f}&-&89.8&71.1/74.2$^*$&70.1&-&-&-&88.6&84.8&86.7&65.1&61.4&63.2&41.9&60.6&49.6\\
      &SABR-T$^\dagger$ \cite{paul2019semantically}&-&88.9&74.0&67.5&-&-&-&\textbf{91.0}&79.7&85.0&\textbf{73.7}&67.2&70.3&41.5&58.8&48.6\\
      &TF-VAEGAN$^\dagger$ \cite{narayan2020latent}&-&92.6&74.7/77.2$^*$&\underline{70.9}&-&-&-&89.6&87.3&88.4&\underline{72.1}&\underline{69.9}&\underline{71.0}&47.1&\underline{62.4}&\underline{53.7}\\
      &GXE\cite{li2019rethinking}&89.8&83.2&61.3&63.5&\textbf{89.0}&\underline{87.7}&\underline{88.4}&{90.0}&80.2&84.8&68.7&57.0&62.3&58.1&45.4&51.0\\
      &LSA$^\dagger$\cite{hanouti2022learning}&-&92.8&\;\;-\;\;/\underline{80.6}$^*$&71.7&-&-&-&86.7&88.5&87.6&-&-&-&\underline{59.5}&46.0&51.8\\
      &SDGN$^\dagger$ \cite{wu2020self}&\underline{92.3}&93.4&74.9&68.4&88.1&87.3&87.7&89.3&\underline{88.8}&\underline{89.1}&70.2&\underline{69.9}&70.1&46.0&62.0&52.8\\
      &STHS-WGAN$^\dagger$\cite{bo2021hardness}&-&\underline{94.9}&\textbf{77.4}&67.5&-&-&-&-&-&-&-&-&-&-&-&-\\\hline
     T &Bi-VAEGAN$^\dagger$(ours) &\textbf{93.9}&\textbf{95.8}&\underline{76.8}/\textbf{82.8}$^*$&\textbf{74.2}&\underline{88.3}&\textbf{89.8}&\textbf{89.1}&\textbf{91.0}&\textbf{90.0}&\textbf{90.4}&71.7&\textbf{71.2}&\textbf{71.5}&45.4&\textbf{66.8}&\textbf{54.1}\\\hline
          \end{tabular}}\\
  \caption{TZSL performance comparison where the unseen class prior is provided when needed.``$\dagger$" denotes the method that adopts the known unseen class prior assumption. `*' denotes the result is obtained using fine-grained  visual descriptions (AK2 in Section \ref{sec:attribute}) for CUB and the competing results marked by `*' are cited from \cite{hanouti2022learning}. \label{tab:2}} 
  \vspace*{-0.3cm}

\end{table*}

\section{EXPERIMENT}

We conduct experiments using four datasets, including  AWA1 \cite{lampert2013attribute}, AWA2 \cite{xian2018zero}, CUB \cite{welinder2010caltech} and  SUN \cite{patterson2012sun}.  Visual features are extracted by the pretrained ResNet-101 \cite{he2016deep}. 
%
Details on the datasets and implementation details are provided in the supplementary material.
We conduct experiments at the feature level following  \cite{xian2018zero,narayan2020latent}. Under the TZSL setup, testing performance of the unseen classes is of interest, for which the average per class top-1 (T1)  accuracy is used, denoted by $\textmd{ACC}^u$ (U).   The ZSL community is also interested in the generalized TZSL performance, i.e., testing performance for both the seen and unseen classes \cite{kong2022compactness,pourpanah2022review,li2022siamese,feng2022non}.  We use the harmonic mean of the average per class top-1 accuracies of the seen and unseen classes to assess it, as  $H = \frac{2 \textmd{ACC}^u\times \textmd{ACC}^s}{\textmd{ACC}^u+ \textmd{ACC}^s}$.
All existing results reported in the tables come from their published papers. When  ``-" appears, such result is missing from the literature.

\subsection{ Result Comparison and  Analysis }

\subsubsection{Known Unseen Class Prior}
We compare performance with both inductive  (I) and transductive (T) state of the arts under the same setting for fair comparison. For TZSL approaches, when class prior of the unseen classes is required, the compared existing techniques assume such information is provided. Therefore,  we first apply the same setting for the proposed Bi-VAEGAN. Table \ref{tab:2} reports the results.  To distinguish from our later results obtained by the proposed prior estimation approach, methods using the provided unseen prior are marked by ``$\dagger$".  Note that we do not report the generalized TZSL performance for STHS-WGAN here. This is because it uses a  harder evaluation setting different from the other approaches by assuming that the unseen and seen data are indistinguishable during training. 

It can be observed from Table \ref{tab:2} that in general, the transductive approaches outperform the inductive approaches with a large gap. The proposed Bi-VAEGAN outperforms the transductive state of the arts, particularly the two baseline frameworks F-VAEGAN and TF-VAEGAN that  Bi-VAEGAN is built on, on almost all the datasets. The new state-of-the-art performance that we have achieved is 93.9\% (AWA1), 95.8\% (AWA2),  and 74.2\% (SUN) for TZSL, while 89.1\% (AWA1), 90.4\% (AWA2), and 54.1\% (SUN) for generalized TZSL. Note that for the CUB dataset that has less intra-class clustering property, we find a simple feature pre-tuning network will further boost the performance from 76.8\% to 78.0\% and we include the discussion in the supplementary material.
It is worth mentioning that  Bi-VAEGAN achieves satisfactory SUN performance where the data is intra-class sample scarce. 
It is challenging to learn from the SUN dataset due to its low sample number of each class that inherently makes the conditional generation less discriminative. 
Bi-VAEGAN provides more discriminative features benefitting from its bi-directional alignment generation 
and the feature augmentation in Section \ref{sec:3.3.4}.
%
\vspace*{-0.2cm}

\begin{table}[th]
  \centering
  \small
   \renewcommand{\multirowsetup}{\centering}
   \begin{tabular}{lcccc}
      \hline
      \multirow{1}{*}{\textbf{Method}}&AWA1&AWA2&CUB&SUN\\\hline
      \multicolumn{5}{l}{\textit{{Non-generative}}} \\

      DSRL\cite{ye2017zero}&74.7&72.8&56.8&48.7\\
      GXE\cite{li2019rethinking}&89.8&\underline{83.2}&61.3&63.5\\
      VSC \cite{wan2019transductive}&- &81.7&71.0&62.2\\\hline
      
      \multicolumn{5}{l}{\textit{{Generative with uniform prior}}} \\
      f-VAEGAN$^\ddagger $ \cite{xian2019f}&62.1&56.5&72.1&69.8\\
      TF-VAEGAN$^\ddagger$\cite{narayan2020latent}&63.0&58.6&\underline{74.5}&71.1\\
      Bi-VAEGAN &66.3&60.3&\textbf{76.8}&\textbf{74.2}\\\hline
      \multicolumn{5}{l}{\textit{Generative with prior estimation}} \\
      Bi-VAEGAN (BBSE) &\underline{90.9}&83.1&72.5&68.4\\
      Bi-VAEGAN (CPE) &\textbf{91.5}&\textbf{85.6}&{74.0}&\underline{71.3}\\\hline
      \end{tabular}\\
      \caption{Performance comparison in $\textmd{ACC}^u$ for both generative and non-generative techniques using different  unseen class priors. `$^\ddagger$' means our reproduced result.\label{tab:3}}
      \vspace*{-0.6cm}
  \end{table}

\subsubsection{Unknown Unseen Class Prior}

\begin{figure}[th]
  \centering
  \begin{subfigure}{0.32\linewidth}
  \centering
  \includegraphics[width=1.1\linewidth]{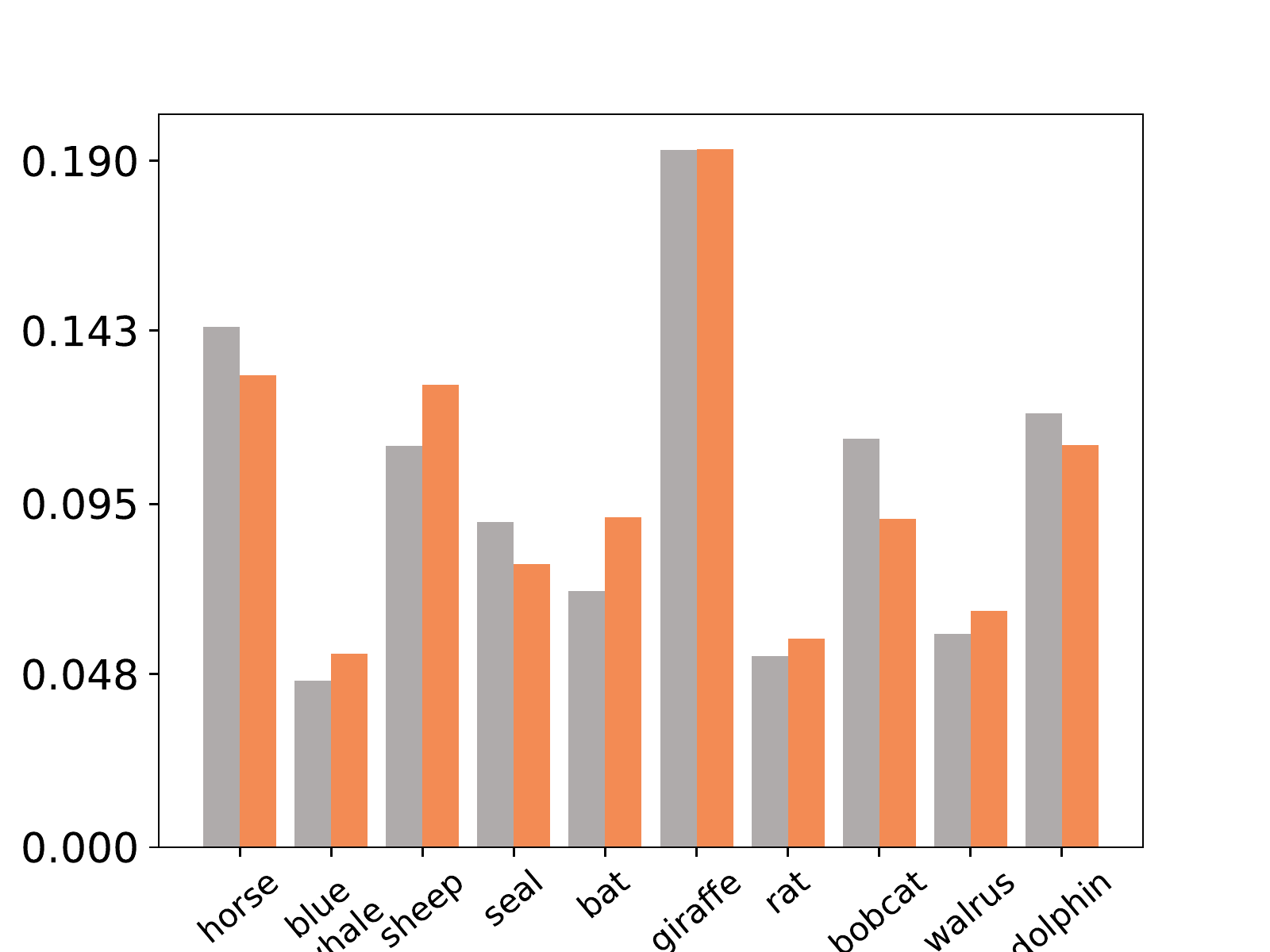}
  \caption{AWA1} 
  \label{fig.5.4}
  \end{subfigure}
  \begin{subfigure}{0.32\linewidth}
  \centering
  \includegraphics[width=1.1\linewidth]{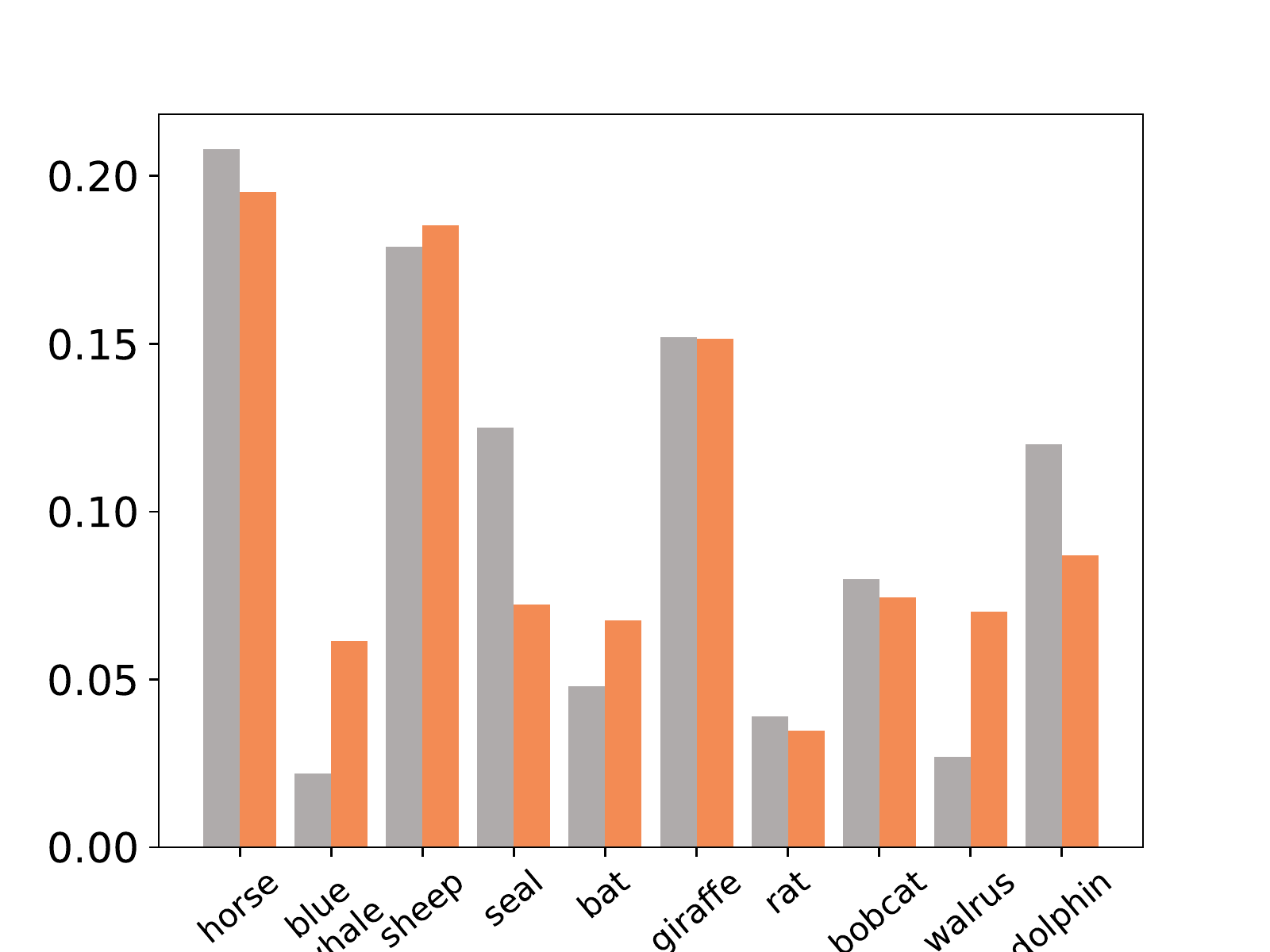}
  \caption{AWA2} 
  \label{fig.5.1}
  \end{subfigure}
  \begin{subfigure}{0.32\linewidth}
      \centering
      \includegraphics[width=1.1\linewidth]{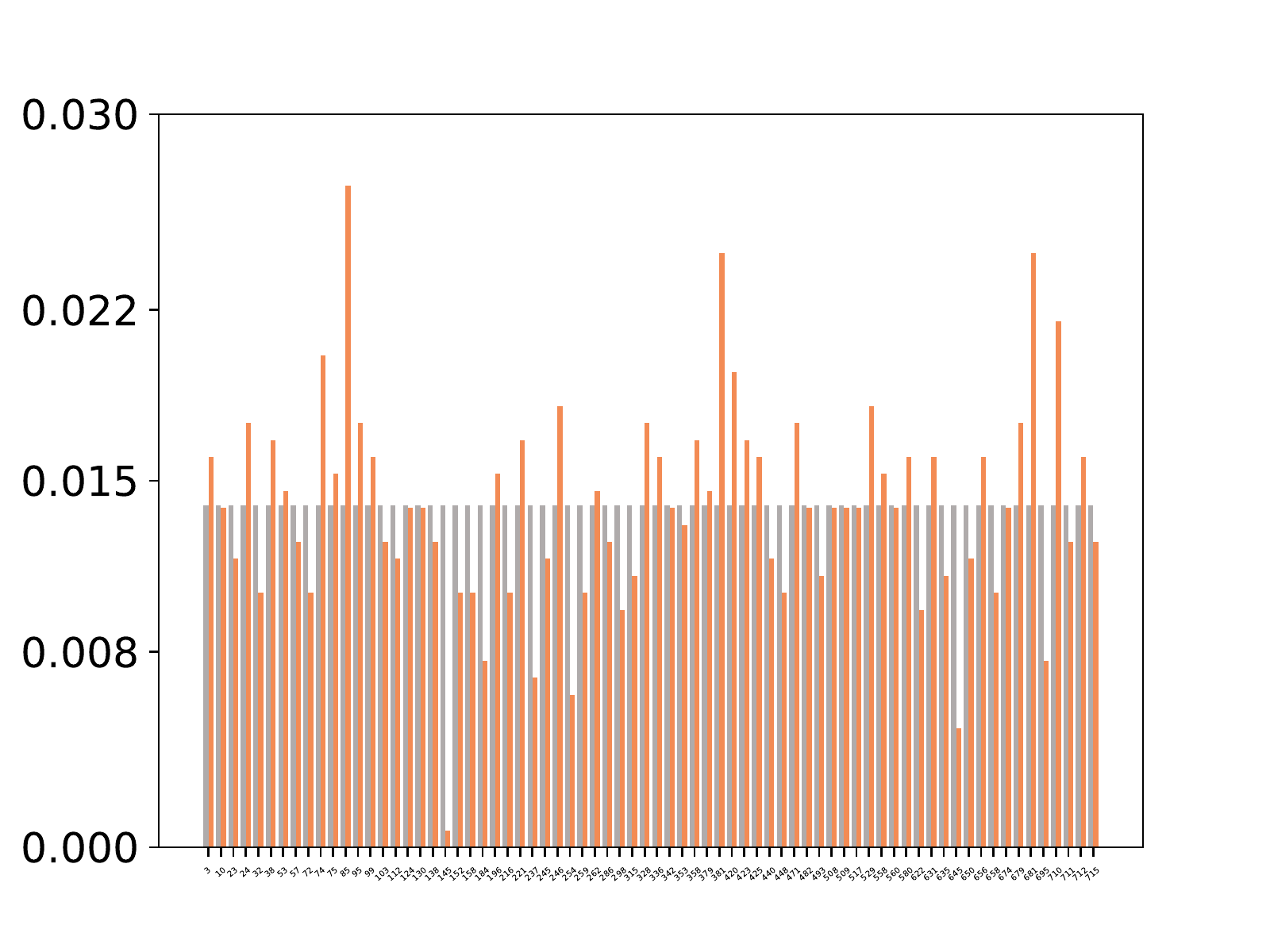}
      \caption{SUN} 
      \label{fig.5.3}
      \end{subfigure}
  \caption{Comparison between the estimated unseen class distribution prior by CPE (orange) and the provided prior computed from the class sizes (gray). \label{fig:att_est}}
  \vspace*{-0.55cm}
\end{figure}

In this experiment, the unseen class prior is not provided. For our method, we compare the proposed prior estimation  with a naive assumption of uniform class prior and a different approach that treats the prior estimation as a label shift problem and solves it by the black box shift estimation (BBSE) method \cite{lipton2018detecting}. BBSE builds upon the strong assumption that $p_G^u(\hat{\bm v}\mid y) = p^u({\bm v}\mid y)$, while our CPE assumes the cluster structure plays an important role in class prior estimation. 
Details on BBSE estimation are provided in the supplementary material.
In Table \ref{tab:3}, we compare our results with the existing ones, where, for methods that need unseen class prior, a uniform prior is used. 
By comparing Table \ref{tab:3} with Table \ref{tab:2} for the generative methods, it can be seen that, when the used unseen class prior differs significantly from the one computed from the real class sizes, there are significant performance drops, e.g.,  over $30\%$ for the extremely unbalanced AWA2 dataset.  
Both BBSE and CPE could provide a satisfactory prior estimation, while our CPE demonstrates consistently better performance.
It can be seen from Figure \ref{fig:att_est} that the CPE prior and the one computed from the real class sizes match pretty well for most classes, and for both the unbalanced and balanced datasets.

\subsection{Ablation Study}

We perform an ablation study to examine the effect of the proposed  $L_2$-feature normalization (FN), 
the  vanilla inductive regressor (R) trained as the level-1 in Eq. (\ref{inductive_training}), 
and transductive regressor (TR) trained adversarially by adding $D^{a}$ as in Eq. (\ref{level1_training}). 
The Min-Max normalized f-VAEGAN is used as an alternative to FN, and  
the inductive regressor that is trained only with the paired seen data is used as an alternative to TR. 
One observation is that FN and TR consistently improve performance over four datasets, respectively. 
We conclude the following from  Table \ref{tab:4}: 
(1) The $L_2$-feature normalization is a free-lunch setting, requiring minimal effort but resulting in a satisfactory performance gain.
(2) A naive implementation of the inductive regressor is beneficial but somehow limited. The regressor trained solely with the seen attributes provides weak constraints to the unseen generation.
(3) The adversarially trained transductive regressor integrates the unseen attribute information and exhibits superiority in the bi-directional synthesis.

  \begin{table}[t]
      \vspace*{-0.1cm}
      \centering
       \renewcommand{\multirowsetup}{\centering}
  
       \begin{tabular}{ccc|ccc}
          \hline
          \multicolumn{3}{c|}{\textbf{Method}}&\multirow{2}{*}{AWA2}&\multirow{2}{*}{CUB}&\multirow{2}{*}{SUN}\\\cline{1-3}
          FN& R&TR&&&\\\hline
          \xmark&\xmark&\xmark&91.6&72.1&69.8\\
          \xmark&\cmark&\xmark&92.3(+0.7)&74.3(+2.1)&70.8(+1.0)\\
          \xmark&\xmark&\cmark&95.5(+3.2)&75.8(+1.5)&72.2(+1.4)\\
          \cmark&\xmark&\cmark&\textbf{95.8}(+0.3)&\textbf{76.8}(+1.0)&\textbf{74.2}(+2.0)\\\hline
          \end{tabular}\\
  
      \caption{Ablation study of transductive ZSL results.}
  \label{tab:4}.
  \vspace*{-0.9cm}
  \end{table}
  
  \begin{figure}[th]
      \vspace*{-0.4cm}
      \centering
      \begin{subfigure}{0.48\linewidth}
          \centering
          \includegraphics[width=1.1\linewidth]{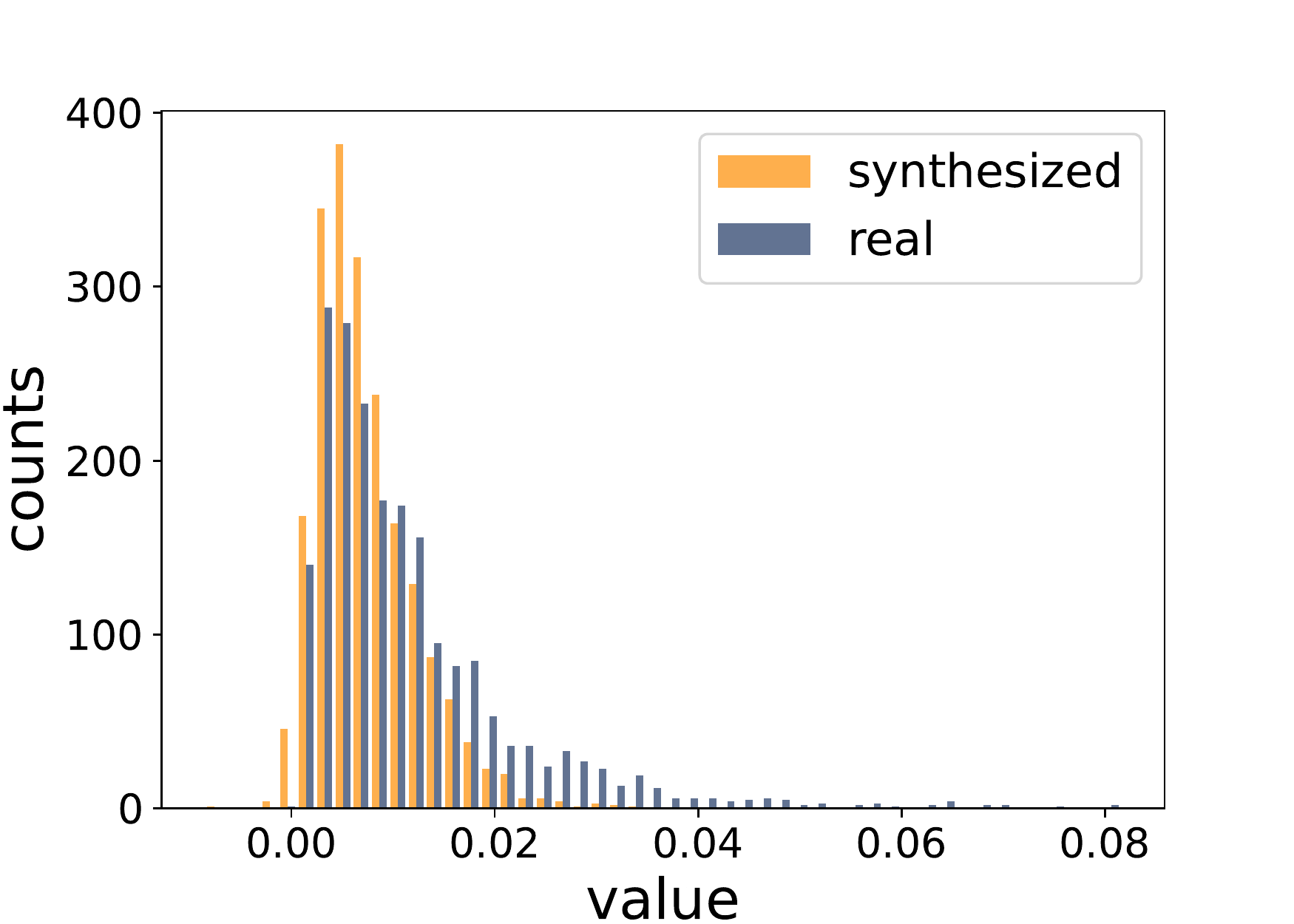}
          \caption{$L_2$ normalization} 
          \label{fig.4.1}
      \end{subfigure}
      \begin{subfigure}{0.48\linewidth}
          \centering
          \includegraphics[width=1.1\linewidth]{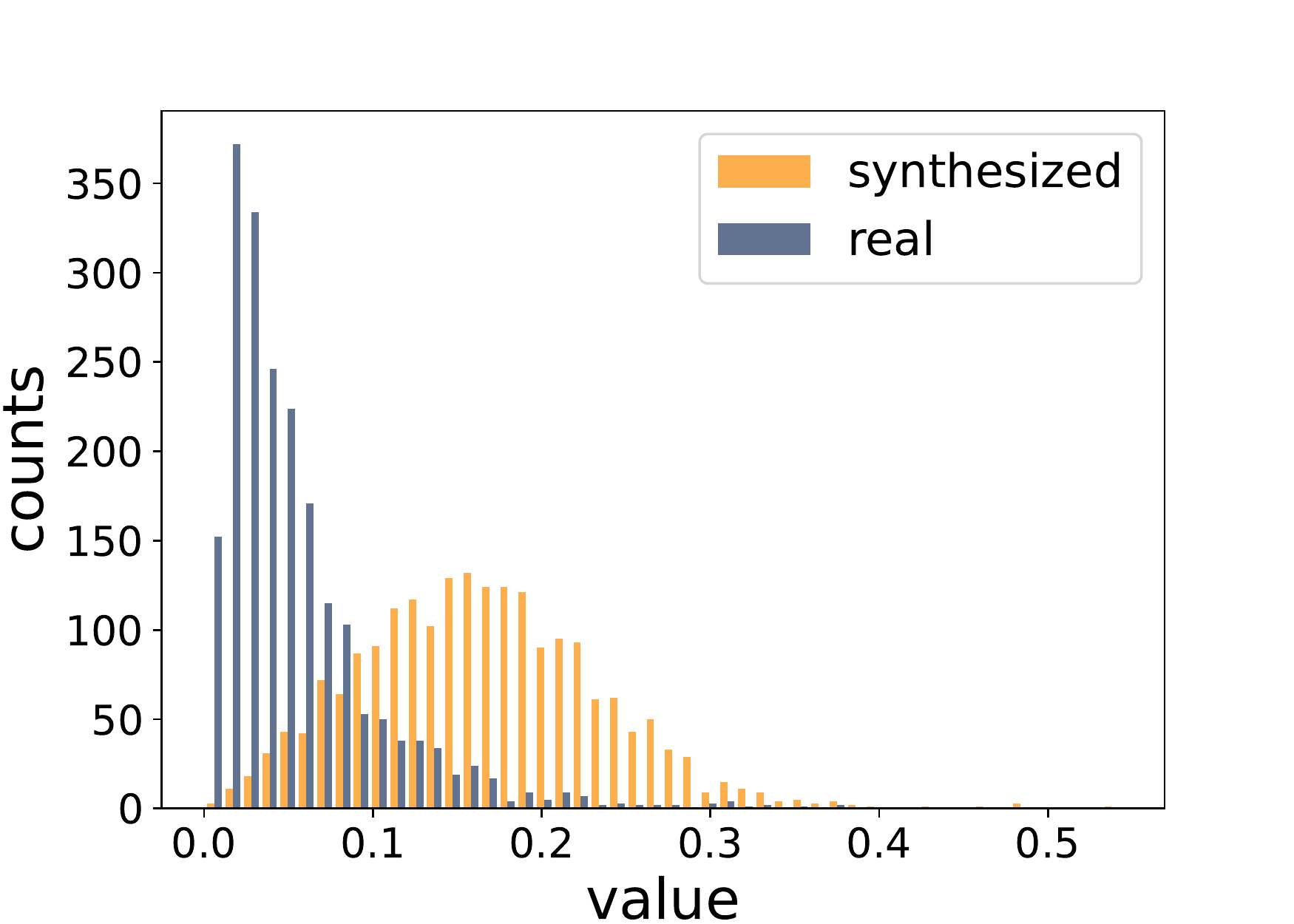}
          \caption{Min-Max normalization} 
          \label{fig.4.2}
      \end{subfigure}
  
      \begin{subfigure}{0.48\linewidth}
          \centering
          \includegraphics[width=1.1\linewidth,height=3cm]{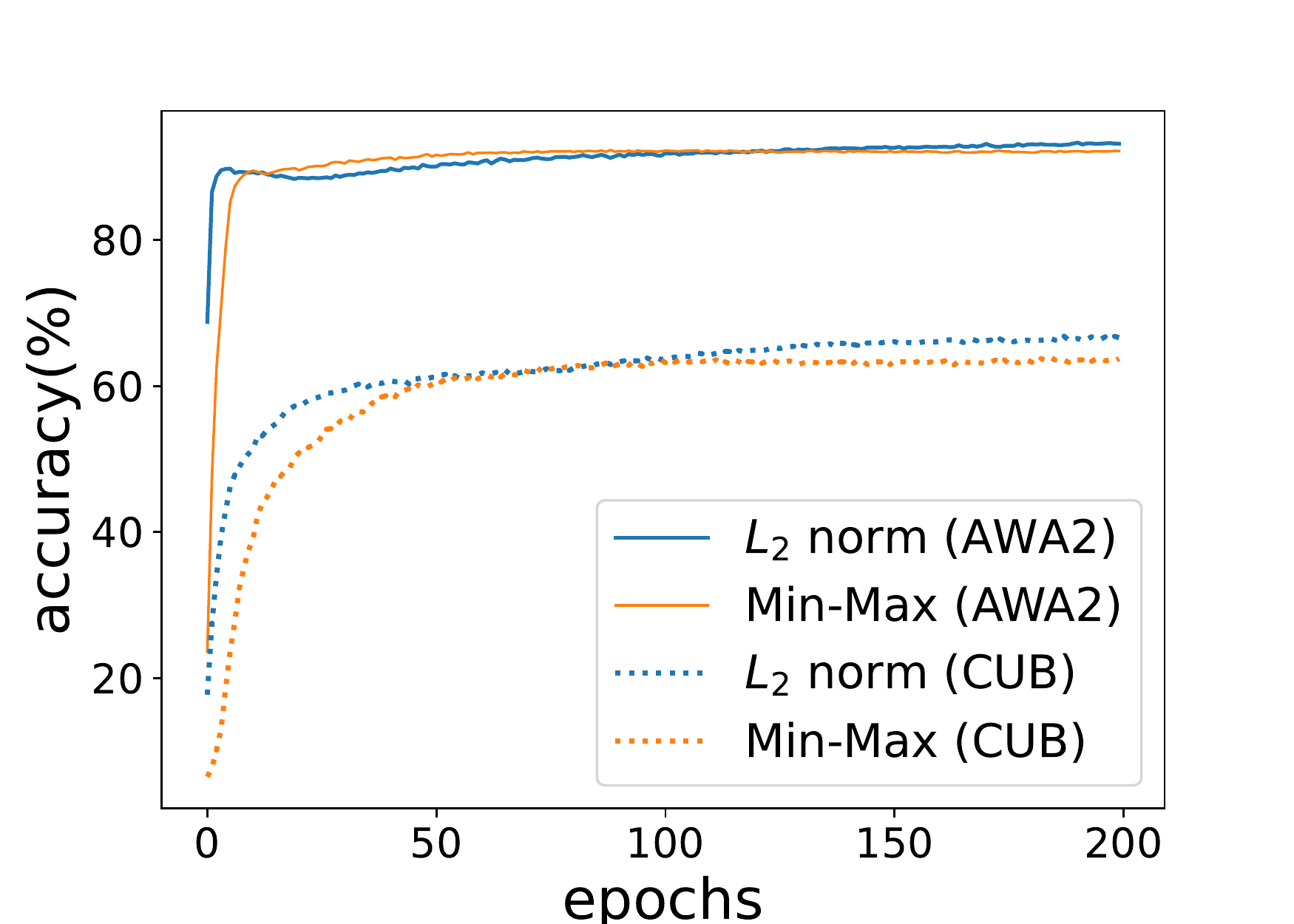}
          \caption{Training accuracy} 
          \label{fig.4.3}
      \end{subfigure} 
      \begin{subfigure}{0.48\linewidth}
          \centering
          \includegraphics[width=1.1\linewidth,height=3cm]{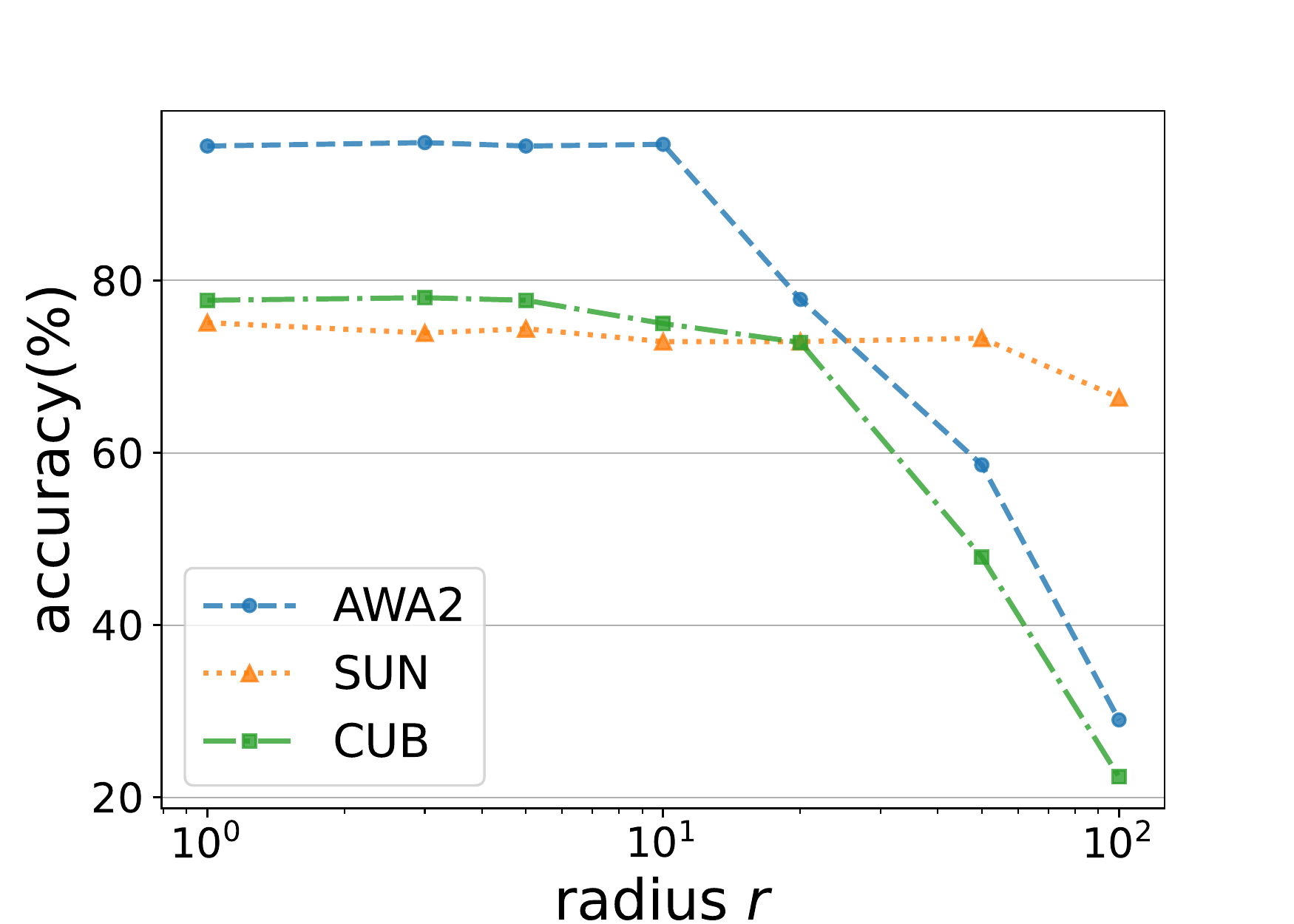}
          \caption{Effect of $r$} 
          \label{fig.4.4}
      \end{subfigure}
      \caption{(a) and (b) compare the real and synthesized feature value distributions of AWA2 after $L_2$ and Min-Max normalizations, respectively. (c) compares the TZSL performance observed during the training for the two normalization approaches using the simplified model on AWA2 and CUB. (d) displays the TZSL performance for different radius values used in $L_2$-normalization. \label{fig:norm}}
      \vspace*{-0.5cm}
      \end{figure}    

\subsection{Further Examinations and Discussion}

\paragraph{On $L_2$-Feature Normalization}
\label{sec:norm}
The difference between using the proposed $L_2$-normalization  and the standard Min-Max normalization is the use of Eq. (\ref{V_norm}) or the sigmoid activation in the last normalization layer of the generator $\bm G$.
To demonstrate the difference between the two approaches, we perform a simple experiment, where the network structure is compressed to contain only three core modules $\bm G$, $D$, and $D^u$. The distributions of the real and synthesized visual features  after two normalizations are compared in Figures \ref{fig.4.1} and \ref{fig.4.2}. 
It can be seen that the $L_2$-normalization results in a better alignment between the two distributions, while Min-Max results in a quite significant gap between the two.
We investigate further the two approaches by looking into their partial derivatives with respect to each dimension, i.e., for the $L_2$ norm, $\frac{d v_i^\prime}{d v_i} =  \frac{r}{\|v_i\|_2}$, and for the Min-Max, $\frac{d \sigma(v_i)}{d v_i}= \sigma(v_i)(1-\sigma(v_i))$.
%
The sigmoid feature has a smaller derivative with a larger input magnitude and vice versa. 
This causes the sigmoid to skew the activated output towards the middle value e.g., 0.5, and this is not suitable when the feature distribution is skewed to one side, especially for those features last activated by ReLU.
It can be seen in Figure \ref{fig.4.3} that   the $L_2$-normalization performs better than Min-Max in terms of a higher accuracy and faster convergence in early training. 
We examine further the effect of the radius parameter $r$ on different datasets in Figure \ref{fig.4.4}. It is observed that a smaller  $r$ could lead to a more stable performance, while a larger $r$ results in an increased gradient that could potentially cause instability in the training. 

\vspace{-0.1cm}

  \begin{figure}[tp]
      \centering  
  
      \begin{subfigure}{0.49\linewidth}
          \centering
          \includegraphics[width=1\linewidth]{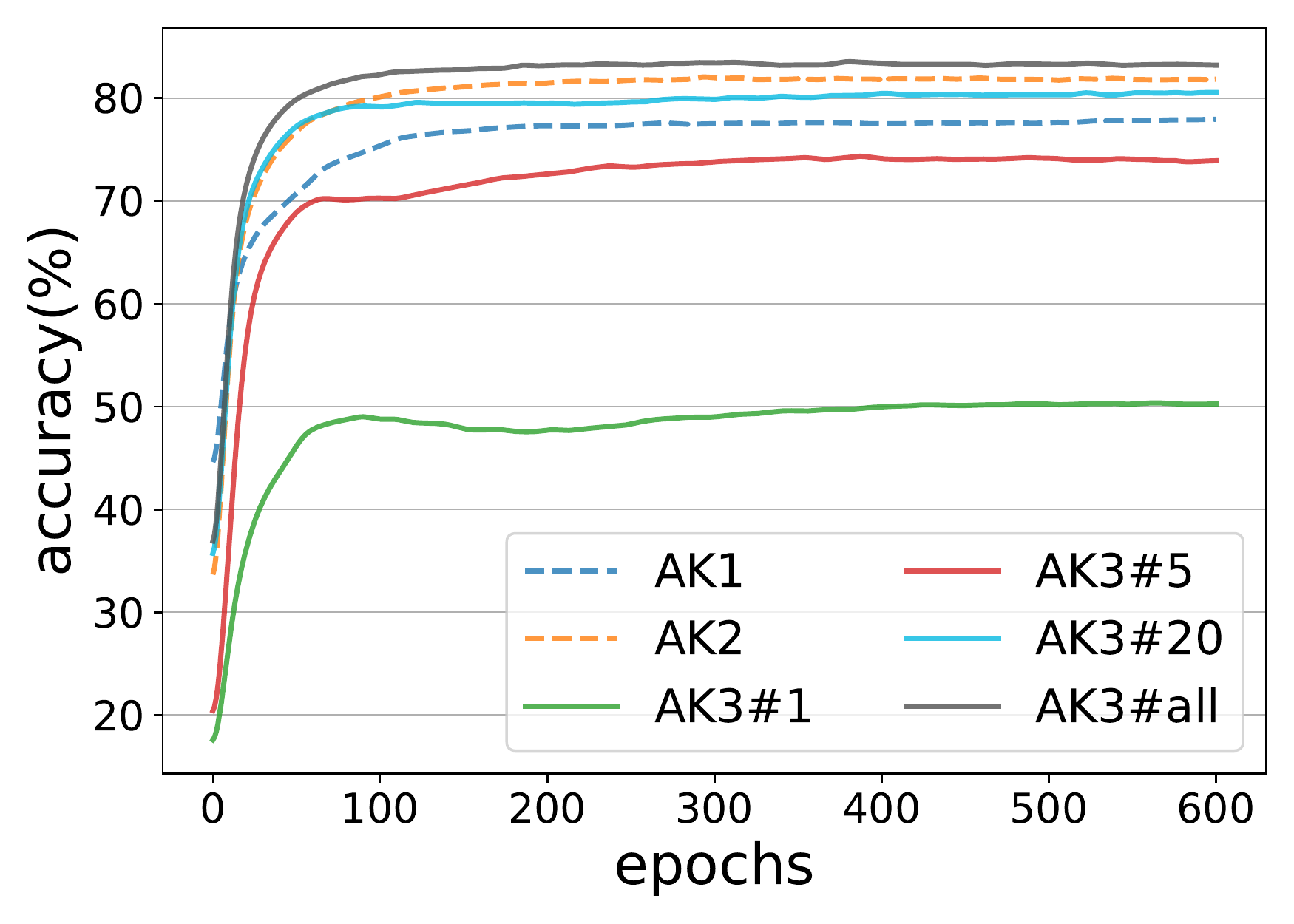}
      \end{subfigure}
      \begin{subfigure}{0.49\linewidth}
          \centering
          \includegraphics[width=1\linewidth]{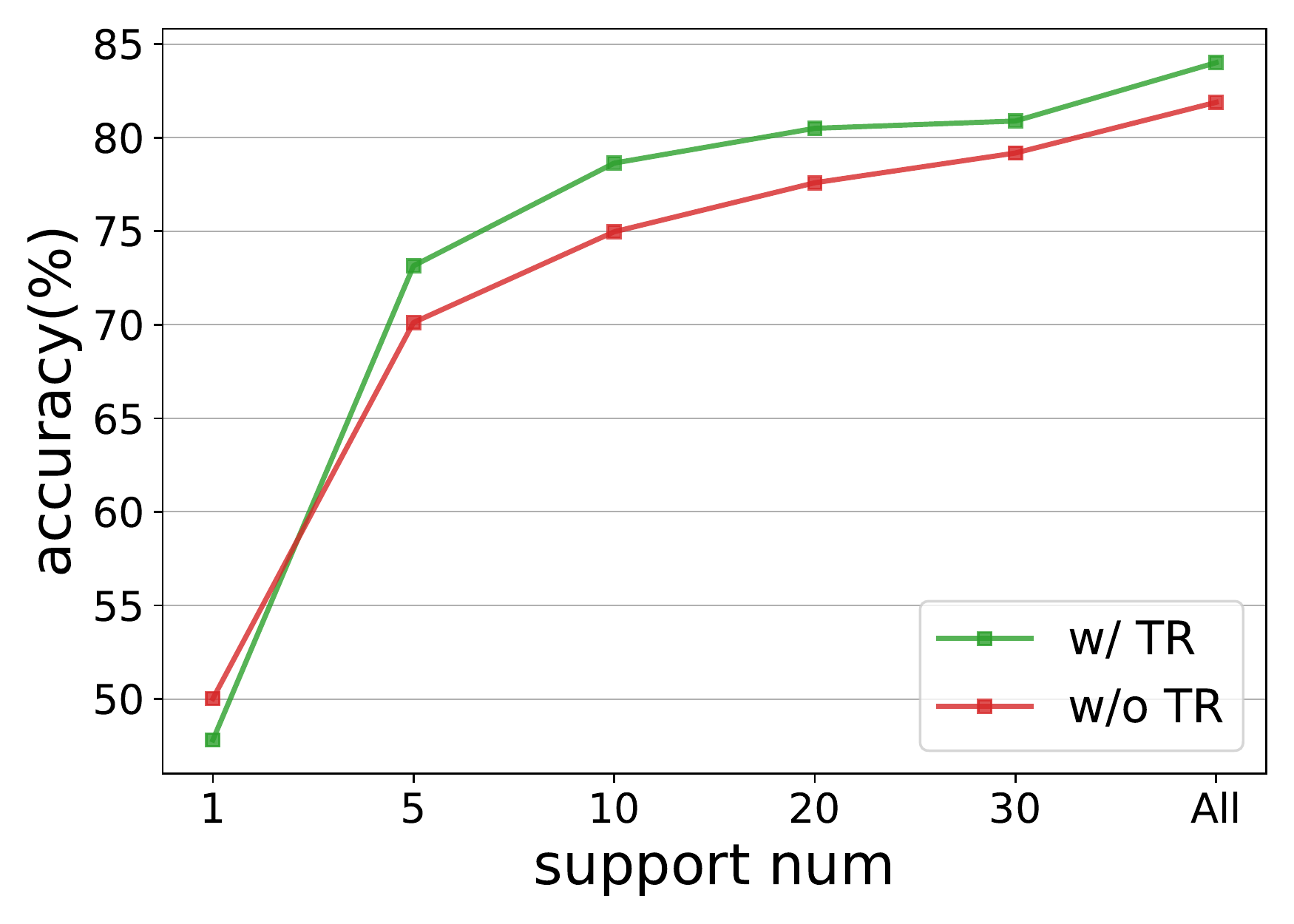}
          \end{subfigure}
      \caption{(a) compares the training accuracies for different auxiliary knowledge. (b) compares TZSL performance with and without the transductive regressor when knowledge AK3 is used.\label{fig:att}}
      \vspace*{-0.58cm}
      \end{figure}

\vspace*{-0.28cm}
     
\paragraph{On Auxiliary Knowledge of CUB} 
\label{sec:attribute}

Auxiliary knowledge plays an overly important role in the success of ZSL, and the motivation of TZSL is to reduce such dependency by learning from unlabelled examples from unseen classes. To examine the effectiveness of our regressor proposed to improve transductive learning and reduce dependency, we conduct experiments using different types of auxiliary knowledge on the CUB dataset.
These include the original 312-dim attribute vectors (AK1), the 1024-dim CNN-RNN embedding from fine-grained visual description \cite{reed2016learning} (AK2), and the 2048-dim averaged visual prototype features of n-shot support (unseen) dataset with label information as assumed in few-shot learning (AK3\#$n$) \cite{li2020adversarial}. 
A ground-truth prototype (AK3\#all) is the strongest auxiliary information where the arrival of a conditional alignment is easier. The effectiveness of the prototypes becomes weaker as the number of support examples decreases, where the outliers tend to dominate the generation. AK2 performs similarly well to the ground-truth prototypes, and this indicates that the embeddings learned from a large-scale language model can serve as a good approximation to the ground-truth visual prototypes. 
The training accuracies are compared in Figure \ref{fig:att}.(a). 
Figure \ref{fig:att}.(b) is produced by removing our  regressor TR (using inductive R) when the model is conditioned on AK3. It is observed that when TR is not employed, a considerable gap opens up for the less informative prototypes. This demonstrates that TR can improve the model's robustness to auxiliary quality. It is important to note that when the auxiliary quality is extremely low (as in the AK3\#1 case), alignment in the unseen domain is hard to realize no matter whether the TR module is presented or not.

\begin{figure}[tp]
  \centering
  \begin{subfigure}{0.55\linewidth}
      \centering
      \includegraphics[width=1.04\linewidth]{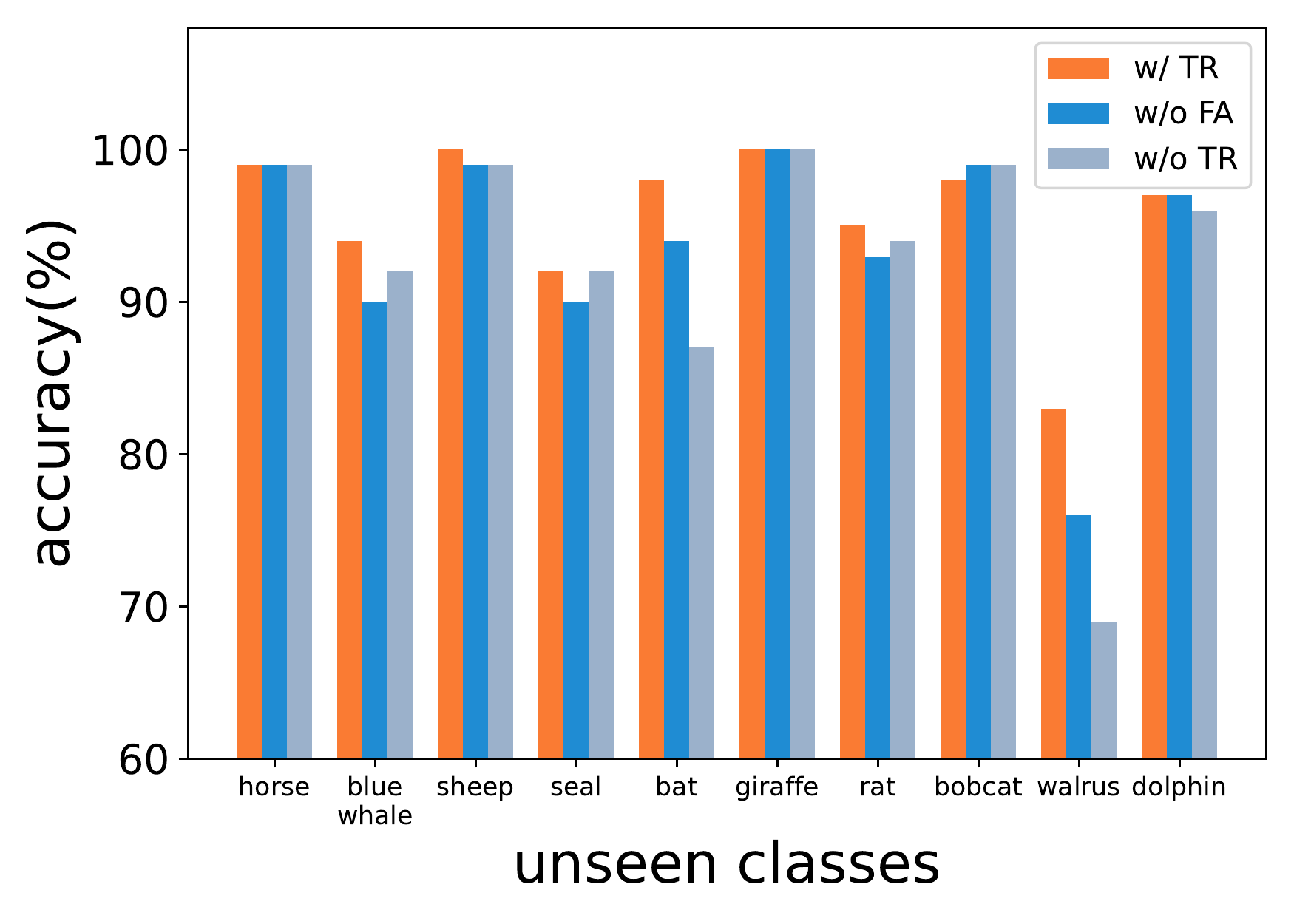}
  \end{subfigure}
  \begin{subfigure}{0.43\linewidth}
      \centering
      \includegraphics[width=1.05\linewidth]{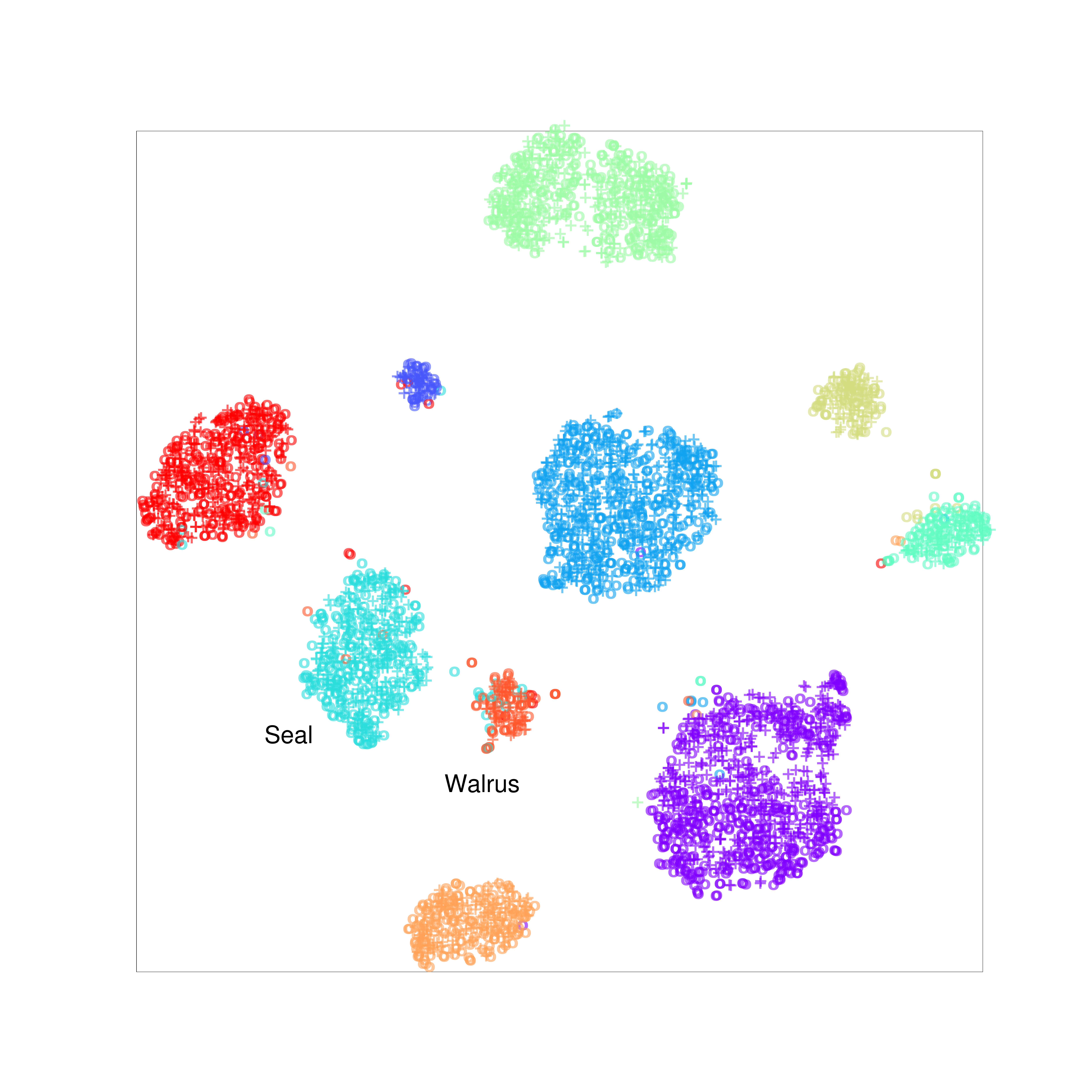}
  \end{subfigure}
  \caption{(a) The classification accuracy of different classes in AWA2. `FA' denotes feature augmentation. (b) T-SNE visualization of augmented real/synthesized feature. \label{fig.fa}}
      \vspace*{-0.45cm}
  \end{figure}

\vspace*{-0.3cm}
\paragraph{On Feature Augmentation} 

The transductive regressor supports distinguishing difficult examples, while the concatenated features based on cross-modal knowledge (see Section \ref{sec:3.3.4}) improve the feature discriminability. Figure \ref{fig.fa}.(a) shows that the regressor leads to a better alignment for the less discriminative classes, such as `bat' and `walrus', and the concatenated features contribute significantly to hardness-aware alignment. Figure \ref{fig.fa}.(b) shows that for the resemble (hard) pairs, such as `walrus' and `seal', the concatenated features are better decoupled in this multi-modal space and the synthesis becomes more discriminative (compare with the visualization in Figure \ref{fig.1}).

\vspace*{-0.17cm}
\section{CONCLUSION}
We have presented a novel bi-directional cross-modal generative model for TZSL. By generating domain-aligned features for the unseen classes from both the forward and backward directions,  the distribution alignment between the visual and auxiliary spaces has been significantly enhanced. By conducting extensive experiments, we have discovered that $L_2$ normalization can result in a more stable training than the commonly used Min-Max normalization. 
We have also conducted a thorough examination of how the unseen class prior can affect the model performance and proposed a more effective prior estimation approach. 
This enables the generative approach to still be robust under the challenging scenario with unknown class priors. 

\vspace*{-0.1cm}
\section{ACKNOWLEDGEMENT}
This work is mainly supported by the National Key Research and Development Program of China (2021ZD0111802). It was also supported in part by the National Key Research and Development Program of China under Grant
2020YFB1406703, and by the National Natural Science Foundation of China (Grants No.62101524 and No.U21B2026).
{\small
\bibliographystyle{ieee_fullname}
\bibliography{egbib}
}
\newpage

\setcounter{section}{0}
\section*{Supplementary Materials}
We present additionally (1) the dataset and implementation details, (2) the explanation of the feature pre-tuning network, (3) an examination of different feature spaces, and (4) the comparison of BBSE and CPE for class prior estimation.

\section{Dataset and Implementation}
\subsection{Dataset} 
We conduct  experiments using four benchmark datasets.
The Animals with Attributes 1\&2 (AWA1 \cite{lampert2013attribute} \& AWA2 \cite{xian2018zero}) contain 30,475\&37,322 samples from a total of 50 classes, and the dimension of the attribute vector is 85.
The Caltech UCSD Bird 200 (CUB) \cite{welinder2010caltech} consists of 11,788 fine-grained images of 200 bird species with an attribute size of 312.
The SUN Scene classification (SUN) \cite{patterson2012sun} dataset has 14,340 samples selectecd  from 717 scenes with an attribute size of 102. More details are shown in Table \ref{tab:S1}.

\begin{figure}[h]
        \centering
        \begin{subfigure}{0.45\linewidth}
        \centering
        \includegraphics[width=1.12\linewidth]{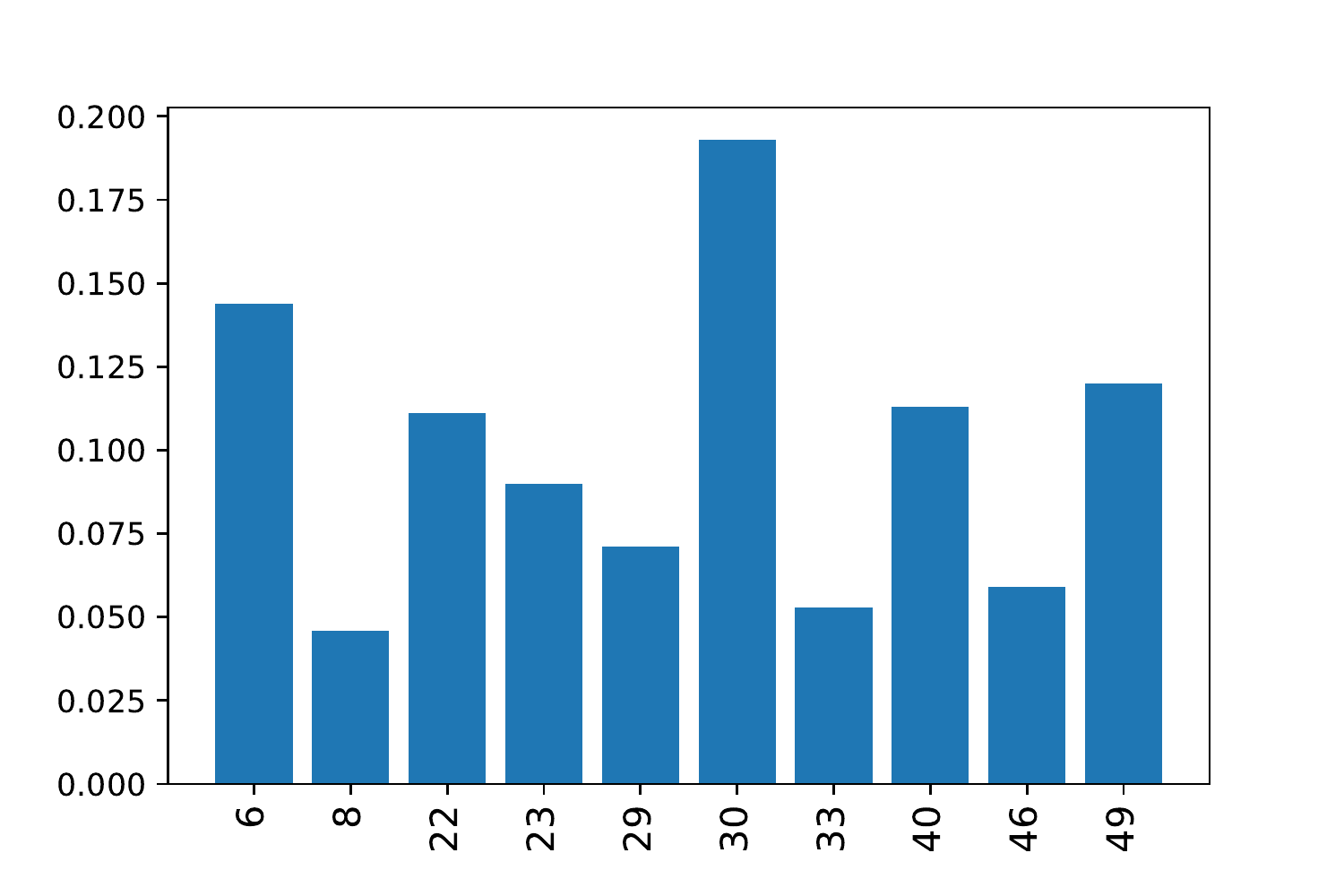}
        \caption{AWA1} 
        \label{fig.3.4}
        \end{subfigure}
        \begin{subfigure}{0.45\linewidth}
        \centering
        \includegraphics[width=1.12\linewidth]{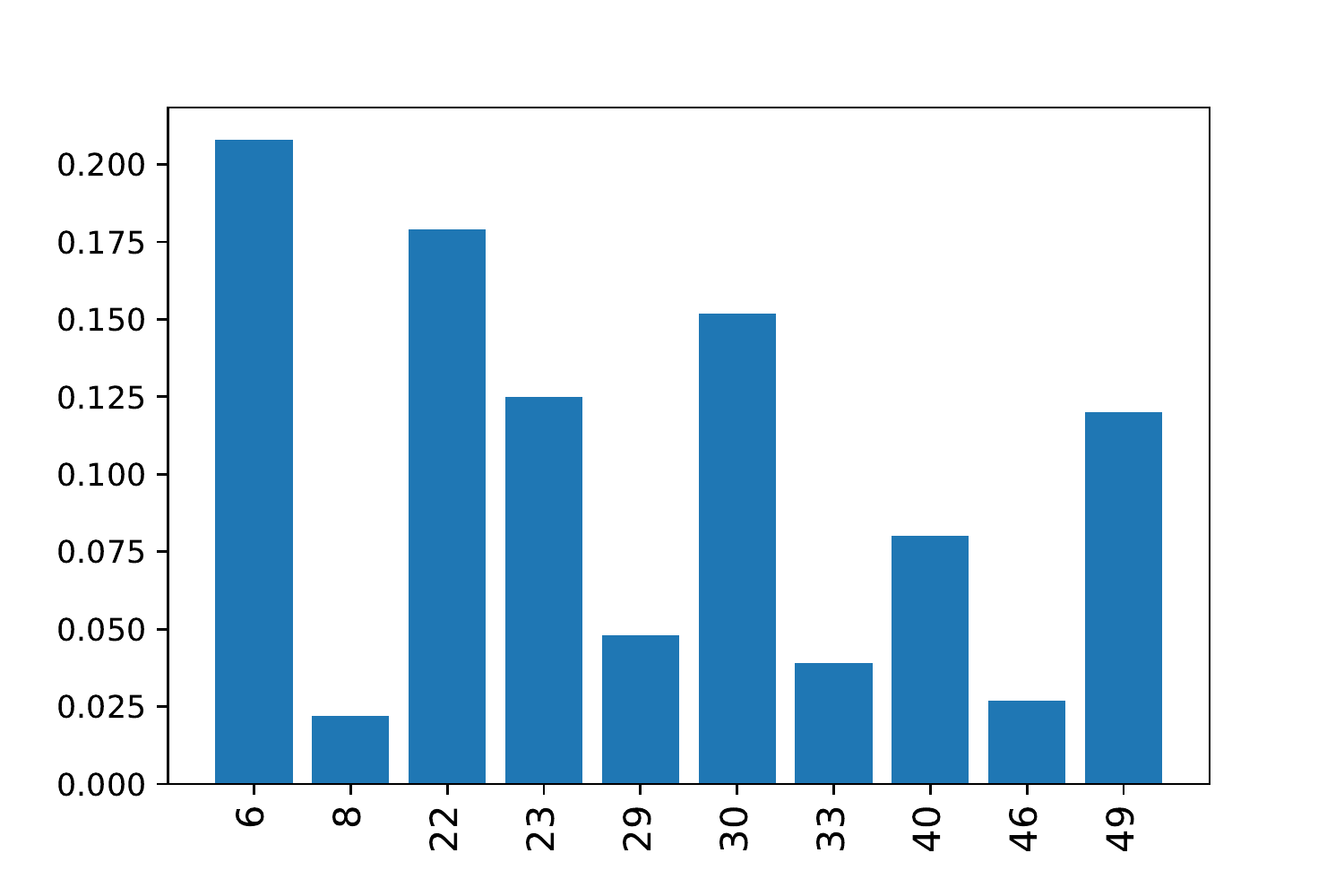}
        \caption{AWA2} 
        \label{fig.3.1}
        \end{subfigure}

        \begin{subfigure}{0.45\linewidth}
        \centering
        \includegraphics[width=1.12\linewidth]{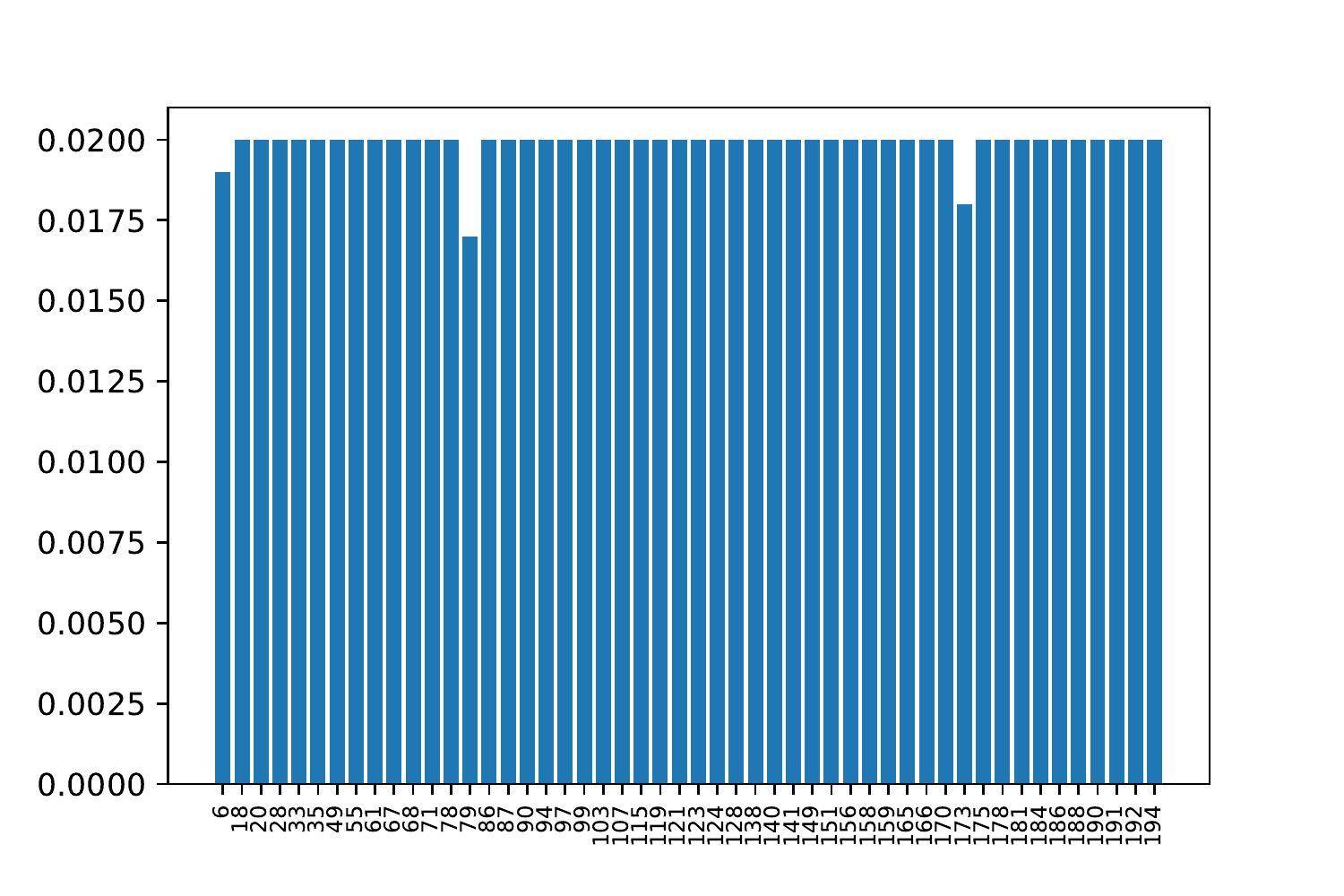}
        \caption{CUB} 
        \label{fig.3.2}
        \end{subfigure}
        \begin{subfigure}{0.45\linewidth}
        \centering
        \includegraphics[width=1.12\linewidth]{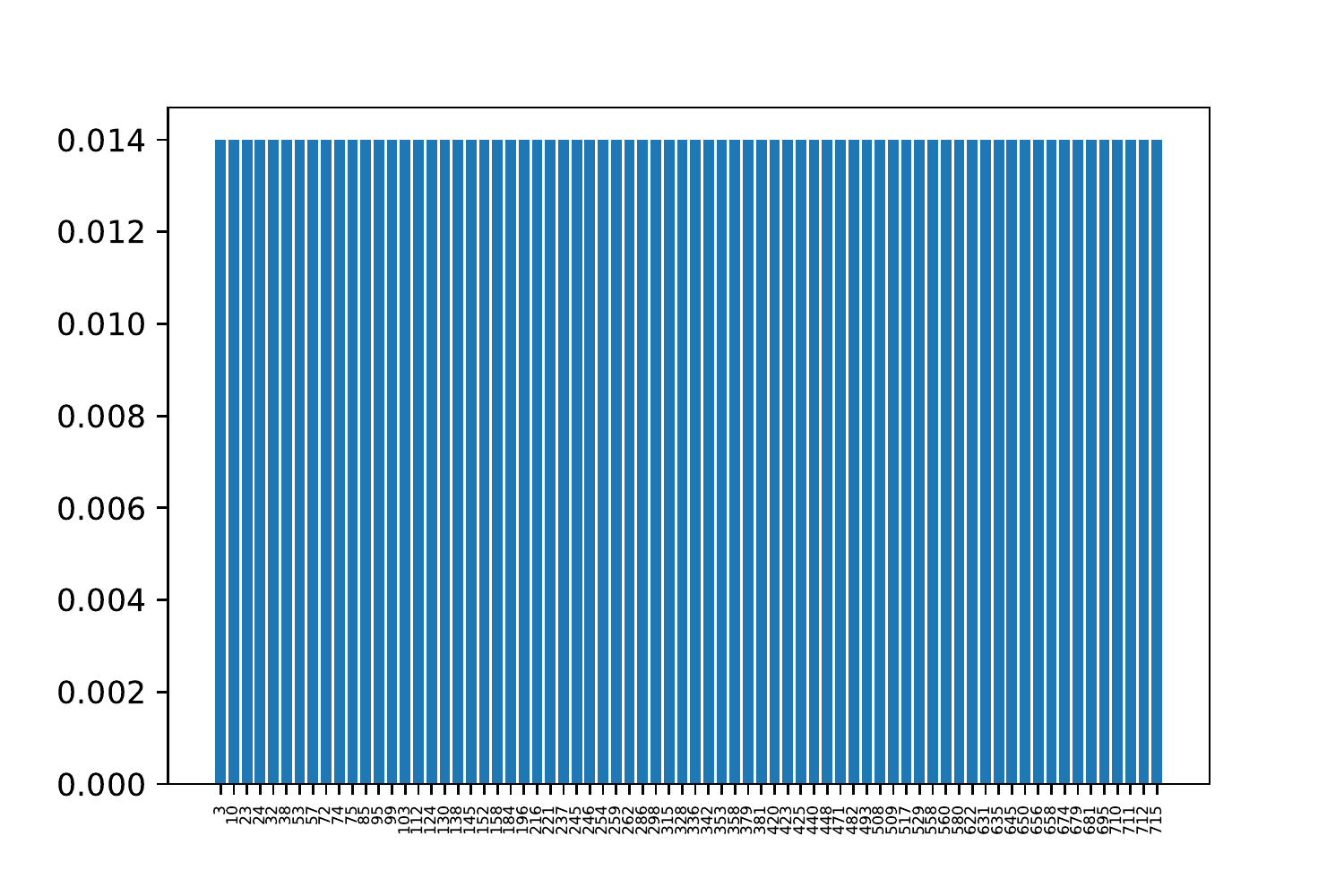}
        \caption{SUN} 
        \label{fig.3.3}
        \end{subfigure}
        \caption{The unseen class prior computed from test data.\label{fig:S1}}
        \end{figure}

\begin{table}[h] 
    \centering
            \setlength{\tabcolsep}{1.8mm}{
            \begin{tabular}{cccccc}\hline
    \textbf{Dataset}&$N$&att.&stc.& $\|\mathcal{Y}^s\|$& $\|\mathcal{Y}^u\|$\\\hline
    AWA1&30,475&85&-&40&10\\
    AWA2&37,322&85&-&40&10\\
    CUB&11,788&312&1,024&150&50\\
    SUN&14,340&102&-&645&72\\
        \hline
            \end{tabular}}
              \caption{Statistics of the four datasets. `att.' denotes the attribute size,  `stc.' is the dimension of semantic information extracted from descriptive sentences \cite{reed2016learning}, $\|\mathcal{Y}^s\|$ and $\|\mathcal{Y}^s\|$ correspond to the numbers of the seen and unseen classes, respectively.}
            \label{tab:S1}
    \end{table}
Figure \ref{fig:S1}  displays the class distribution prior  estimated from the  class information of the testing samples from the unseen classes, i.e., the percentage of the samples  contained by each class, for the four datasets. 
AWA1 and AWA2 have unbalanced class priors, while CUB and SUN have class priors close to a uniform distribution.
AWA2 has  more samples from those popular classes like `horse' and `dolphin'. 
  
\subsection{Implementation} 
In the training of  all our modules, we use AdamW optimizer \cite{DBLP:conf/iclr/LoshchilovH19} with a learning rate of 0.001 and ($\beta_1$, $\beta_2)$ is set as (0.5, 0.999). The encoder $\bm E$, decoder $\bm G$ and regressor $\bm R$ in Bi-VAEGAN are all two-layer MLPs, in which the hidden layer output has 4,096 dimensions and the inner activation layer is LeakyReLU. 
The conditional visual critic $D$, unconditional visual critic $D^u$, and attribute critic $D^a$ are two-layer MLPs where the output of the last layer is a scalar, and the WGAN gradient penalty coefficient is set as 10. 
The  level-1 and level-2 trainings proceed alternatively. We conduct one-step level-1  training  for every five steps of level-2 training to accelerate the training speed.
The training epochs for AWA1, AWA2, CUB, and SUN are set to be 300, 300, 600, and 400, respectively. In the inference stage, the synthesized feature number of each class  are set to be 3000, 3000, 400, and 400 for for AWA1, AWA2, CUB, and SUN, respectively. The classifier $f$ is a single fully connected (FC) layer and its output dimension is equal to the number of unseen classes for TZSL or the number of both seen and unseen classes for generalized TZSL. 

The used hyper-parameters for reporting results are $r$=1 for $L_2$ normalization, $\lambda$=1, $\alpha$=1, $\beta$=10 and $\gamma$=10, where the setting of $\alpha$, $\gamma$ and the WGAN critic training are the same as TF-VAEGAN \cite{narayan2020latent}. 
In level-1 training, $\lambda$ is less sensitive and thus set to 1. 
Values of $r$ and $\beta$ are searched within $\{1,2,5,\cdots,100\}$ and $\{0.01,0.1,1,10,100\}$, respectively.  Due to the unavailability of a test split in the datasets, we report our results on the validation split, consistent with previous works \cite{narayan2020latent,wu2020self}. For conventional Zero-Shot Learning (ZSL) and Generalized Zero-Shot Learning (GZSL), a more rigorous setting is desired, especially under the impact of current large models.


\section{Feature Pre-tuning Network}

\label{sec:pre-tune}
\begin{figure}[t]
    \centering
    \includegraphics[width=1\linewidth]{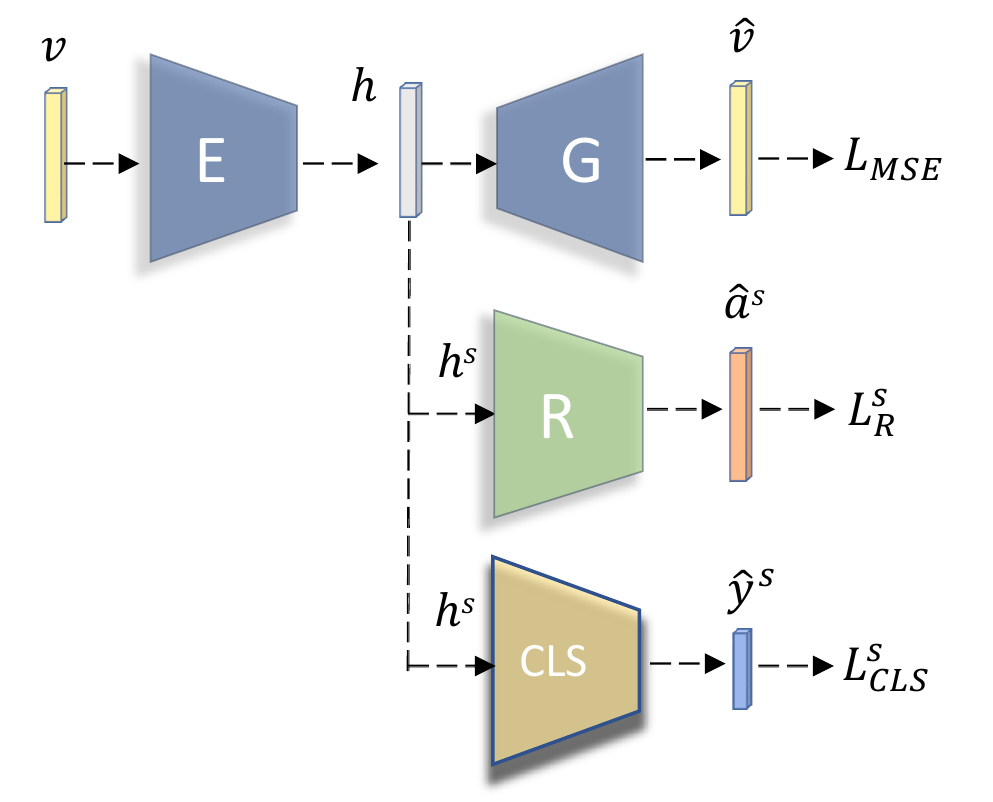}
    \caption{The used feature pre-tuning network.}
    \label{fig:pre-tune}
\end{figure}

\subsection{Used Approach}

For the CUB dataset, we pre-tune the pre-trained features using a supervised neural network  of which the architecture is shown in Figure \ref{fig:pre-tune}.
It builds on an auto-encoder network ($\bm E^\prime$ and $\bm G^\prime$) and consists of two supervised modules that work in the latent space, acting as a regressor ($\bm R^\prime$) and a classifier (CLS). `$^\prime$' denotes it is a different module from the one in the main text.
The input and latent  features share the same feature dimension, i.e., 2,048 for the pre-trained ResNet-101.
Only the seen classes receive supervision from the two supervised modules. The  training  objective for feature pre-tuning is,
\begin{align}
    \begin{split}
        \operatorname*{min}_{\bm E^\prime, \bm G^\prime, \bm R^\prime, CLS}  L_{MSE} + L_{\bm R^\prime}^s + L_{CLS}^s,
    \end{split}
\end{align}
where 
\begin{align}
L_{CLS}^s (\mathcal{V}^s) = \mathbb{E} [ \operatorname{log}(P(y|v^s)) ].
\end{align}
The latent features are extracted by the encoder $\bm E^\prime$ after training for 15 epochs for both the seen and unseen classes. These replace the original visual features to be used as the input of Bi-VAEGAN.

\subsection{Result Comparison}

\begin{figure}[t]
    \centering
    \begin{subfigure}{0.48\linewidth}
        \centering
        \includegraphics[width=1.12\linewidth]{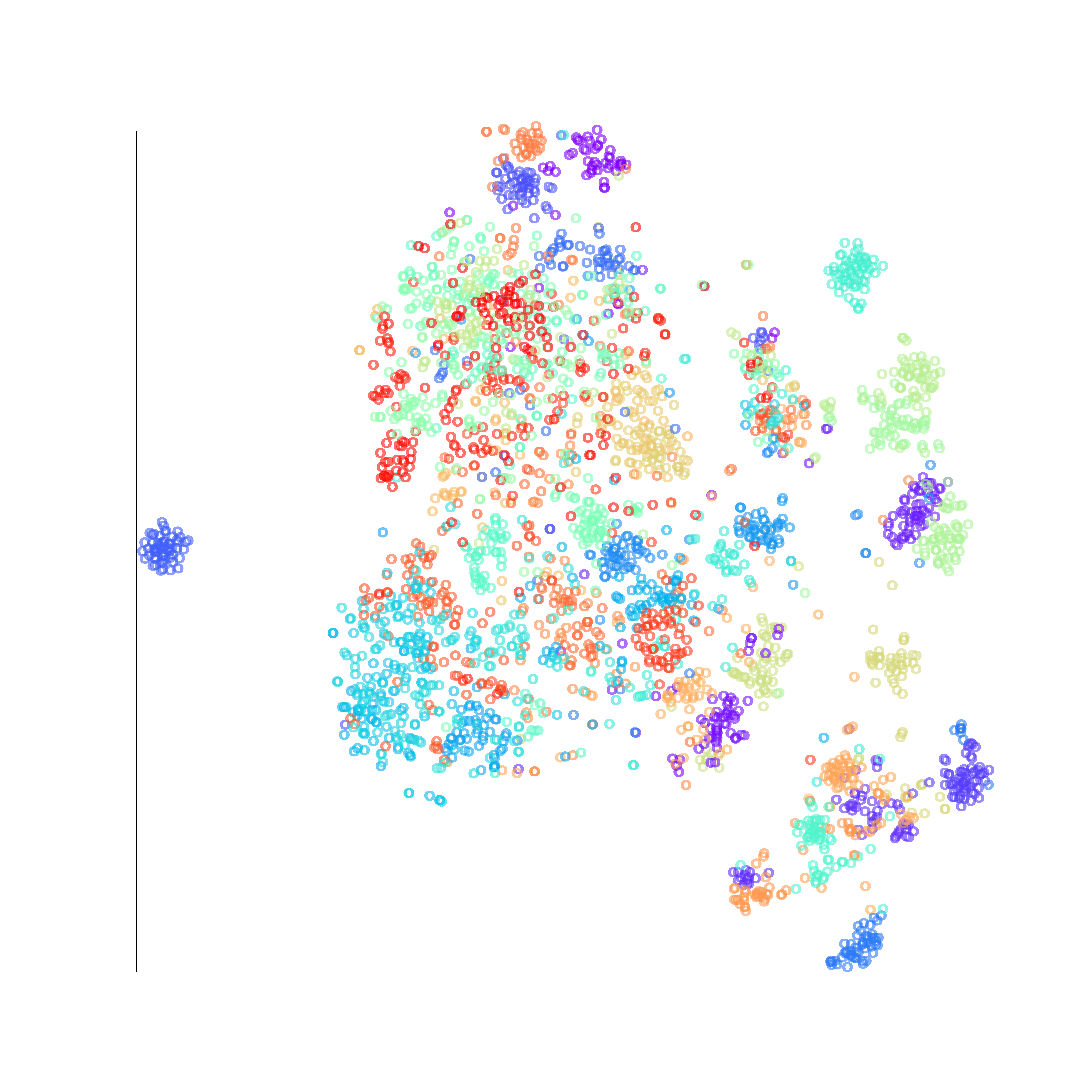}
        \caption{untuned (CUB)} 
    \end{subfigure}
    \begin{subfigure}{0.48\linewidth}
    \centering
    \includegraphics[width=1.12\linewidth]{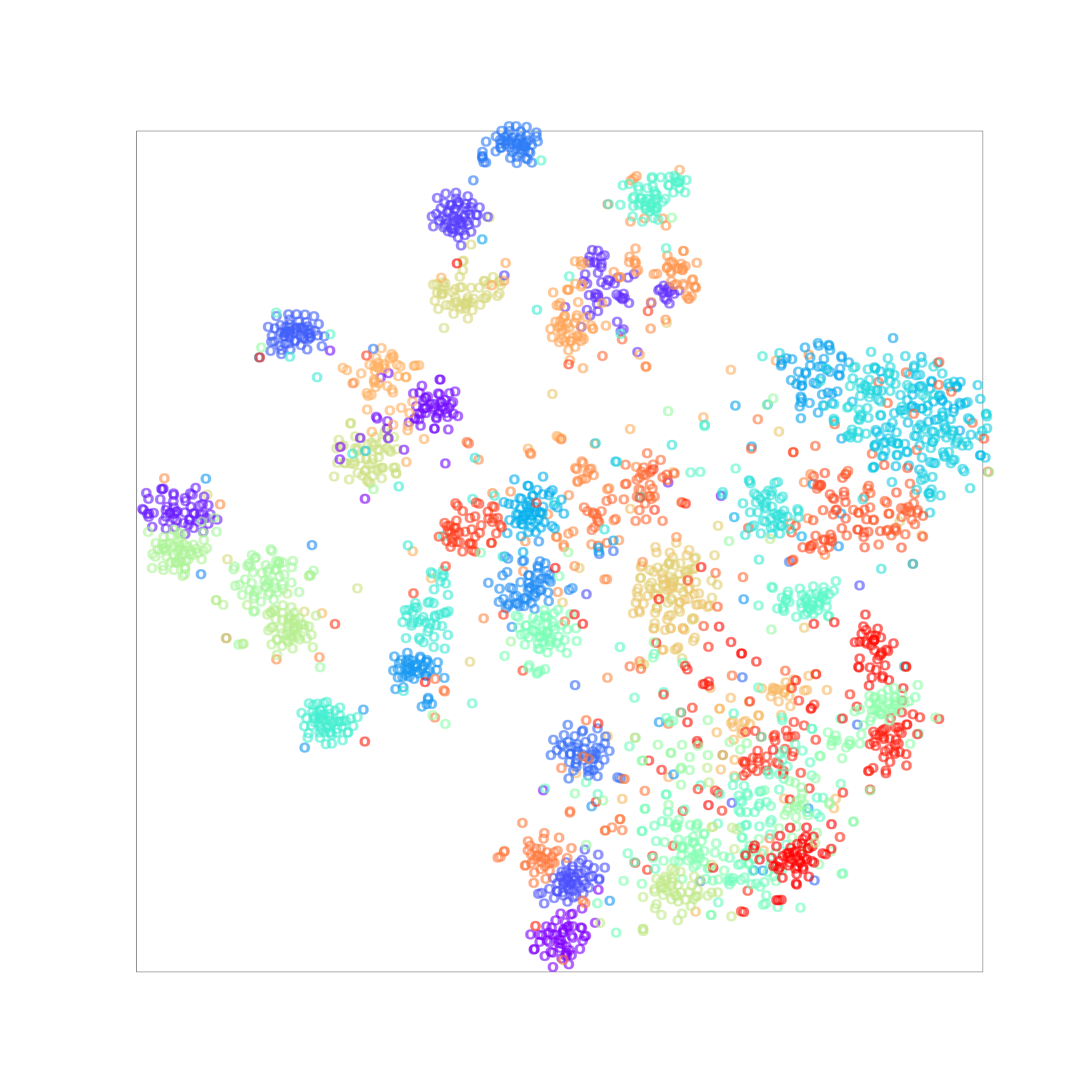}
    \caption{pre-tuned (CUB)} 
    \end{subfigure}
    \caption{Visualization of vanilla and pre-tuned CUB features.\label{tune.cub}}
    \end{figure}

\begin{figure}[t]
    \centering
    \begin{subfigure}{0.48\linewidth}
        \centering
        \includegraphics[width=1.12\linewidth]{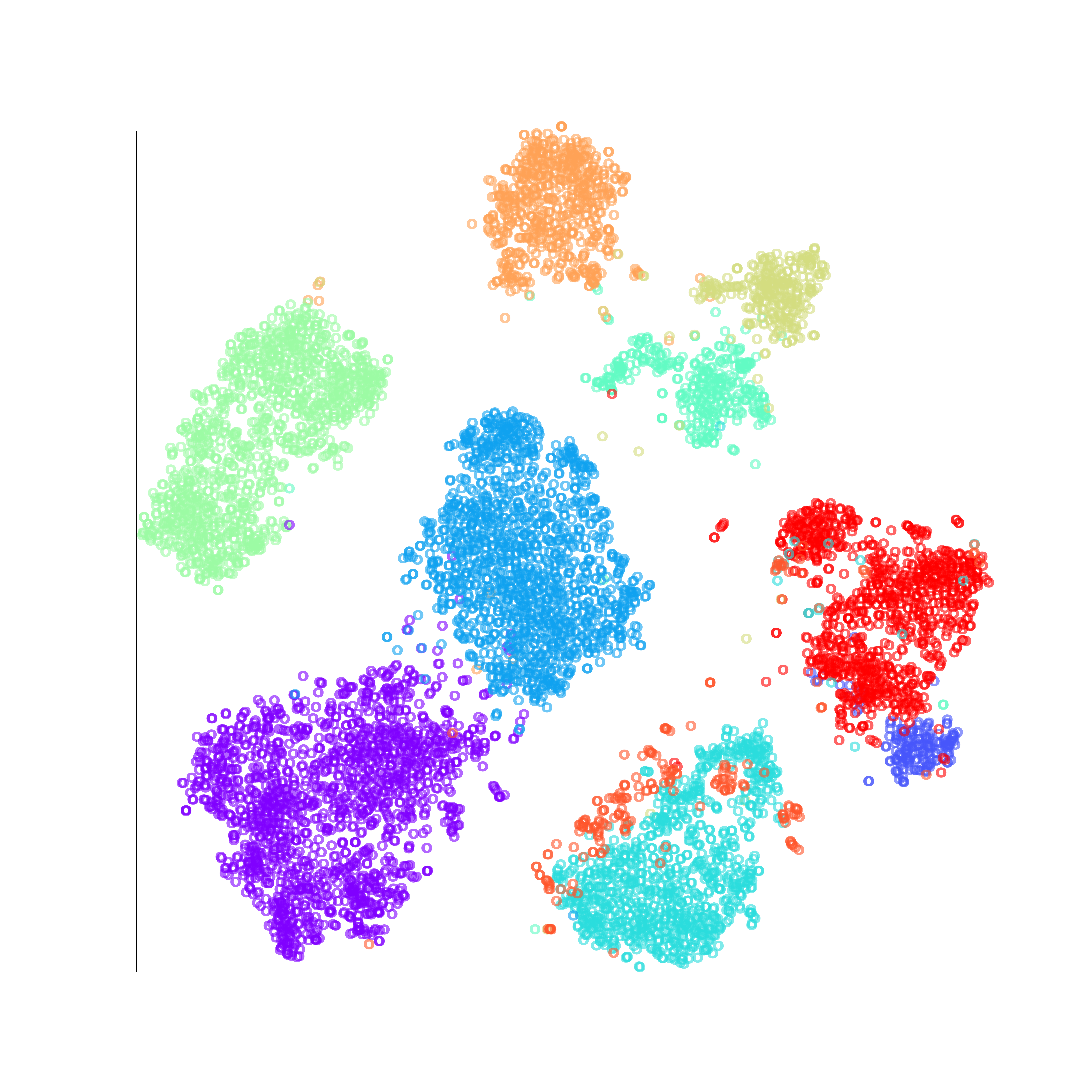}
        \caption{untuned (AWA2)} 
        \label{fig.10.1}
    \end{subfigure}
    \begin{subfigure}{0.48\linewidth}
    \centering
    \includegraphics[width=1.12\linewidth]{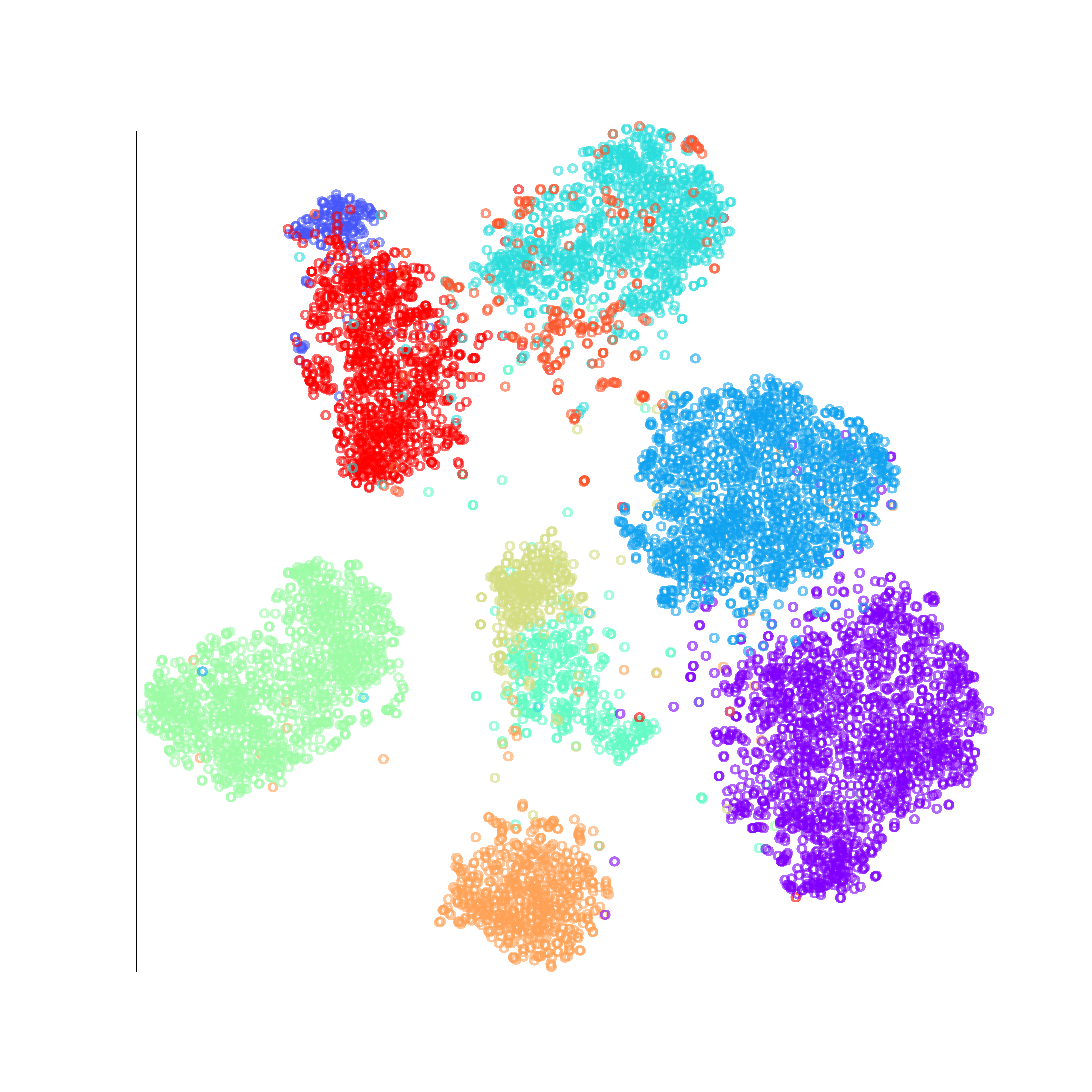}
    \caption{pre-tuned (AWA2)} 
    \label{fig.10.2}
    \end{subfigure}
    \caption{Visualization of vanilla and pre-tuned AWA2 features.\label{tune.awa2}}
    \end{figure}

Figures \ref{tune.cub} and \ref{tune.awa2} visualize the tuned and untuned features for the CUB and AWA2 datasets, 
using the visulization tool t-SNE? .
The tuned features exhibit  more clear cluster structure for the cross-domain dataset CUB. 
%
%
It should be noted that our feature pre-tuning network will not be beneficial for datasets that already have a satisfactory cluster structure, and somehow the cluster property could be damaged.

    \begin{table}[b]
        \centering 
        \small
        \begin{tabular}{cccc}
            \textbf{Model} & AWA2 & CUB$^{\text{AK1}}$& CUB$^{\text{AK2}}$ \\\hline
            \multicolumn{4}{l}{\textit{w/o pre-tuning}}\\
    
            Simple-GAN & 92.7   &68.1 & 79.8\\ 
            Bi-VAEGAN &  \textbf{95.8}   &76.8 &\textbf{82.8}\\\hline
            \multicolumn{4}{l}{\textit{w/ pre-tuning}}\\
            Simple-GAN &  88.9 ($-$3.8)  &76.9 ($+$8.8) & 80.3 ($+$0.5)\\ 
            Bi-VAEGAN & 90.0 ($-$5.8)   &\textbf{78.0} ($+$1.2) &82.0 ($-$0.8)\\\hline
        \end{tabular}
        \caption[]{The effect of feature pre-tuning on AWA2 and CUB. Simple-GAN is a simplified version of  Bi-VAEGAN. 
      All performances are shown in percentage ($\%$).
      CUB$^{\text{AK1}}$ conditions on the original attribute information (AK1) while CUB$^\text{AK2}$ conditions on the semantic embedding (AK2) extracted from the fine-grained visual description. \label{tab:S2}}
    \end{table}

Table \ref{tab:S2} demonstrates the effect of feature pre-tuning on AWA2 and CUB datasets. We name the simplified model that only contains $\bm G$, $\bm D$, and $\bm D^u$ as a Simple-GAN. 
Both Simple-GAN and Bi-VAEGAN use $L_2$ feature normalization. A key observation is that for CUB, feature pre-tuning introduces a noticeable improvement for both models, i.e., $+$8.8 and $+$1.2 respectively, when using the less informative AK1 knowledge.
Notably, Simple-GAN significantly benefits from this straightforward strategy and performs comparably  to the untuned Bi-VAEGAN, e.g., 76.9\% vs. 76.8\%.
This shows that despite the fact that no additional supervision (regressor) is applied, the visual feature alignment for the tuned features is substantially simpler.
We could conclude that the tuned features can lead to a better inter-class discriminability, which enables an easier alignment between the auxiliary and visual spaces  when the class distribution prior is known.

Another observation is that  Simple-GAN  benefits less from the feature pre-tuneing ($+0.5$) when it conditions on the more informative AK2.
Bi-VAEGAN also shows a small performance drop ($-$0.8) with the feature pre-tuning. 
We could conclude that the pre-tuned features are less effective when the auxiliary information is already strong enough.
Besides, for the AWA2 dataset, pre-tuning decreases the inter-class discriminability as shown in Figure \ref{tune.awa2}, and a significant performance drop ($-$3.8, $-$5.8) is observed.
These indicate that feature pre-tuning is not a completely free-lunch approach and that cross-domain datasets may benefit more from it.
Transductive regressor could also achieve a competitive knowledge transfer for the cross-domain dataset.
It is easier to provide a better alignment since it doest not change the original features extracted from the powerful backbone. 
Overall, both the transductive regressor method and the feature pre-tuning offer advantages of their own and may complement one another in complex real-world  circumstances.


  

\section{Feature Augmentation}
As a bi-directional distribution alignment technique for TZSL, our Bi-VAEGAN allows the regressor and generator to independently solve the TZSL problem.
In the inference phase, we compare the performance of using four different feature spaces, i.e., attribute space $\mathcal{A}$, hidden space $\mathcal{H} \in \mathbb{R}^{4096}$  corresponding to the hidden representation of the regressor, visual space $\mathcal{V}$ and the augmented multi-modal space $\mathcal{A\times H\times V}$.
To conduct inference on $\mathcal{A}$, we have two straightforward choices:
(1) Use only the transductively trained $\bm R$ and infer for the test unseen data $\bm{R}(V^u)$ using a 1-nearest neighbor  (1-NN) classifier.
(2) Use both $\bm {G}$ and $\bm {R}$, synthesize the labeled unseen set $\langle \bm{R}(\hat{V}^u_{\bm G}),\hat{Y}^u_{\bm G} \rangle$ in attribute space,  train a neural network classifier using the  labeled set that includes the synthesized examples and infer for $\bm{R}(V^u)$ using this classifier. A similar method of inference can also be applied to the hidden space when this option is chosen.

{\bf Discussion.} 
Table \ref{tab:S3} shows the TZSL top-1 accuracy on three datasets using different spaces to conduct inference.
The observation could be summarized as, 
(1) $\bm{R}$ could be served as an individual module to conduct TZSL inference, but it is much less discriminative that $\bm{G}$.
(2) When using $\bm{G}$ to conduct inference,  a multi-modal space is preferred and the rank of spaces' discriminability is  $\mathcal{H}>\mathcal{V}>\mathcal{A}$. We attribute the hidden space absorbing the knowledge of both transductive generator and regressor and the larger dimensionality is also preferred to alleviate the hubness problem.

\begin{table}[h]   

\setlength\tabcolsep{5pt}
    \centering 
    \small
    \begin{tabular}{cccccc}
       \textbf{Module} &\textbf{Space} & AWA2 & CUB$^{\text{AK1}}$  & CUB$^{\text{AK2}}$ & SUN \\\hline
$\bm{R}$ & $\mathcal{A}$ & 73.2  &64.5  &45.0 &52.6\\\hline 
$\bm{G}$ & $\mathcal{V}$ &  94.2  &75.0 &81.8&71.8\\\hline
\multirow{3}{*}{$\bm{R, G}$}       &$\mathcal{A}$ & 89.8  & 65.6 &67.3&53.2\\ 
       & $\mathcal{H}$ &  \textbf{95.8}   &\textbf{77.2} &82.7&73.8\\
       & $\mathcal{A\times H\times V}$ &  \textbf{95.8}  &76.8 &\textbf{82.8}&\textbf{74.2}\\\hline

    \end{tabular}
    \caption[]{TZSL results of Bi-VAEGAN using different feature spaces.\label{tab:S3}}
\end{table}

\section{BBSE vs. CPE for Class Prior Estimation}

\begin{figure}[thp]
    \centering
    \begin{subfigure}{0.45\linewidth}
        \centering
        \includegraphics[width=1.12\linewidth]{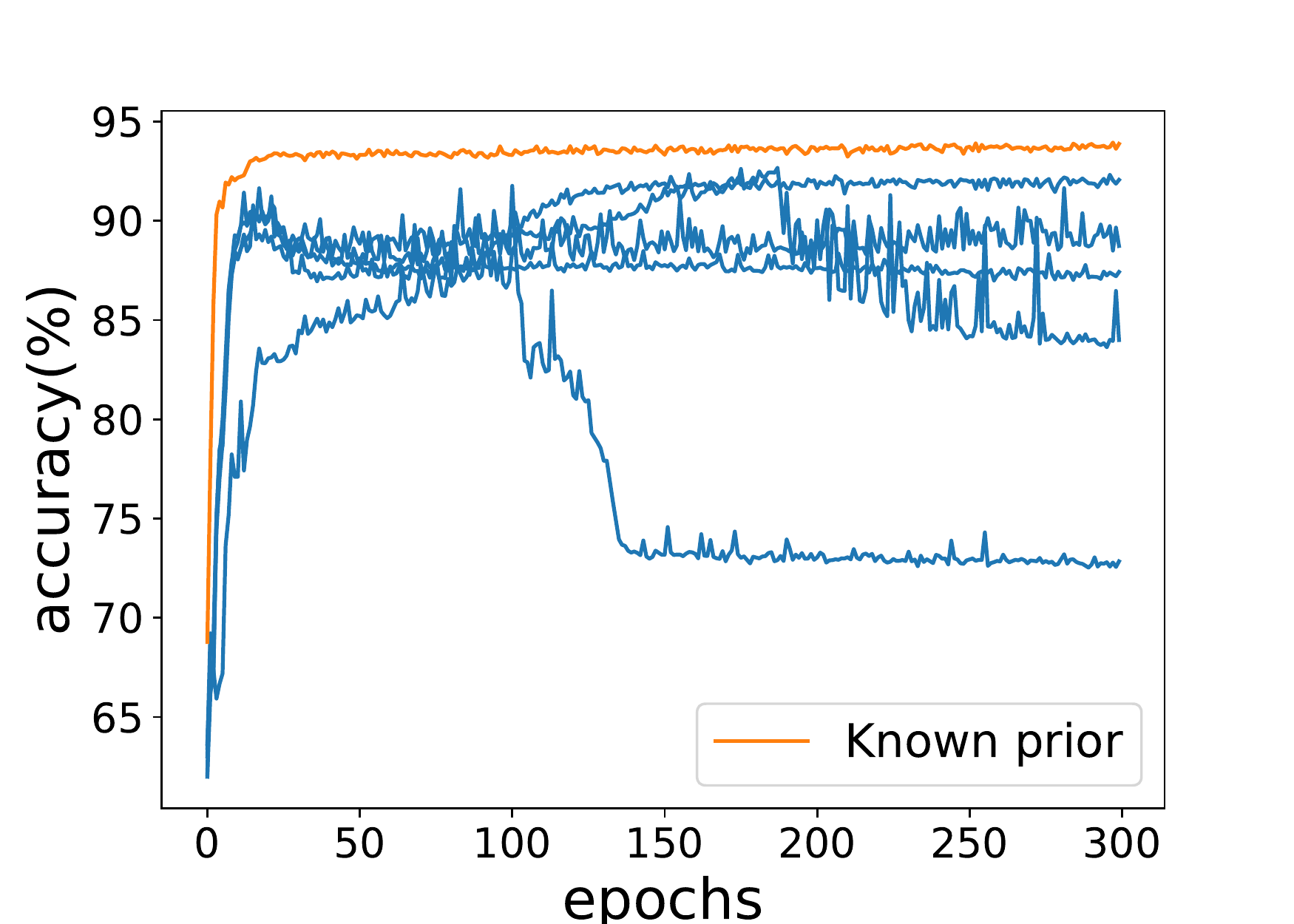}
        \caption{AWA1 (BBSE)} 
        \label{fig.S3.1}
    \end{subfigure}
    \begin{subfigure}{0.45\linewidth}
    \centering
    \includegraphics[width=1.12\linewidth]{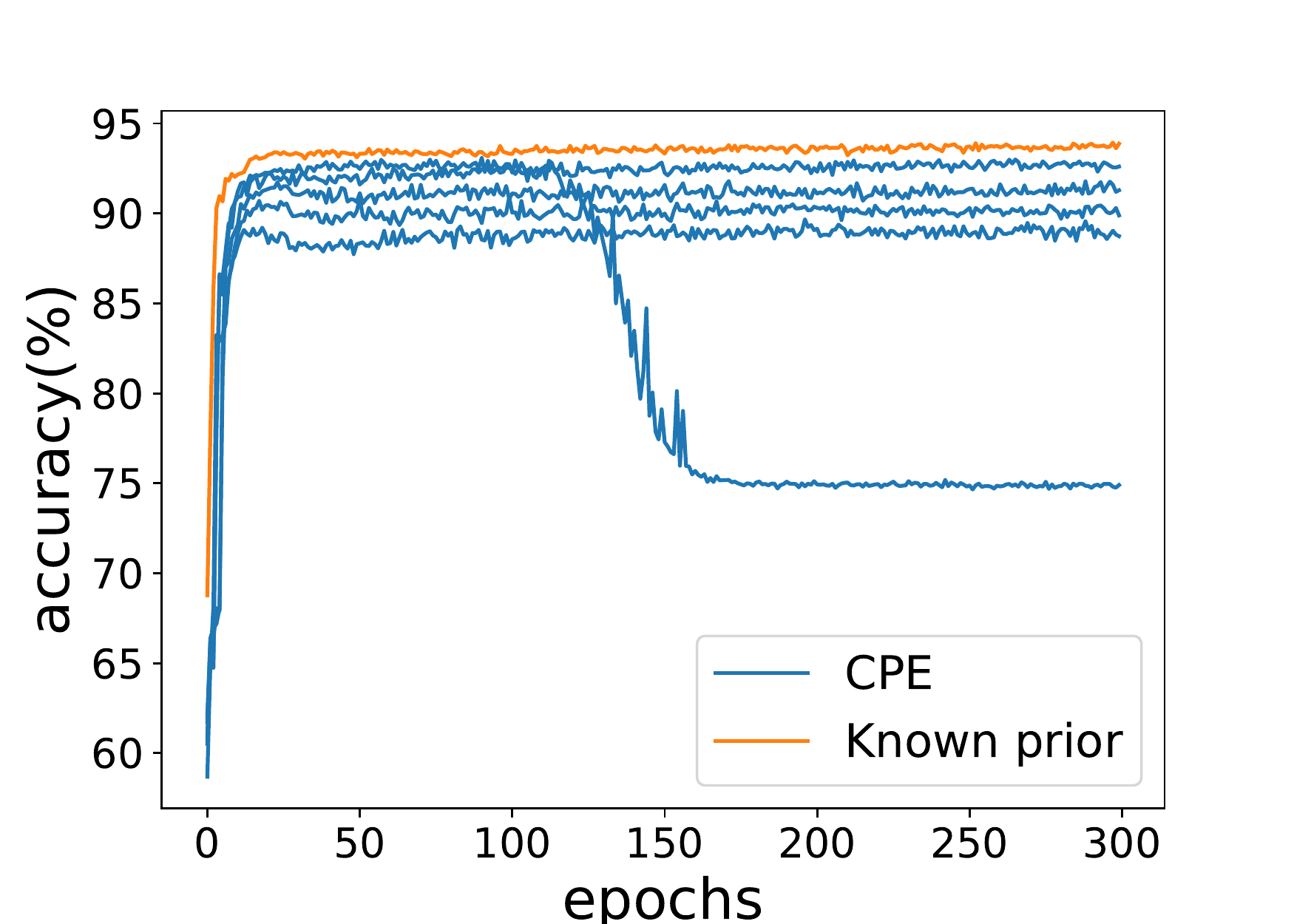}
    \caption{AWA1 (CPE)} 
    \label{fig.S3.2}
    \end{subfigure}%

    \begin{subfigure}{0.45\linewidth}
        \centering
        \includegraphics[width=1.12\linewidth]{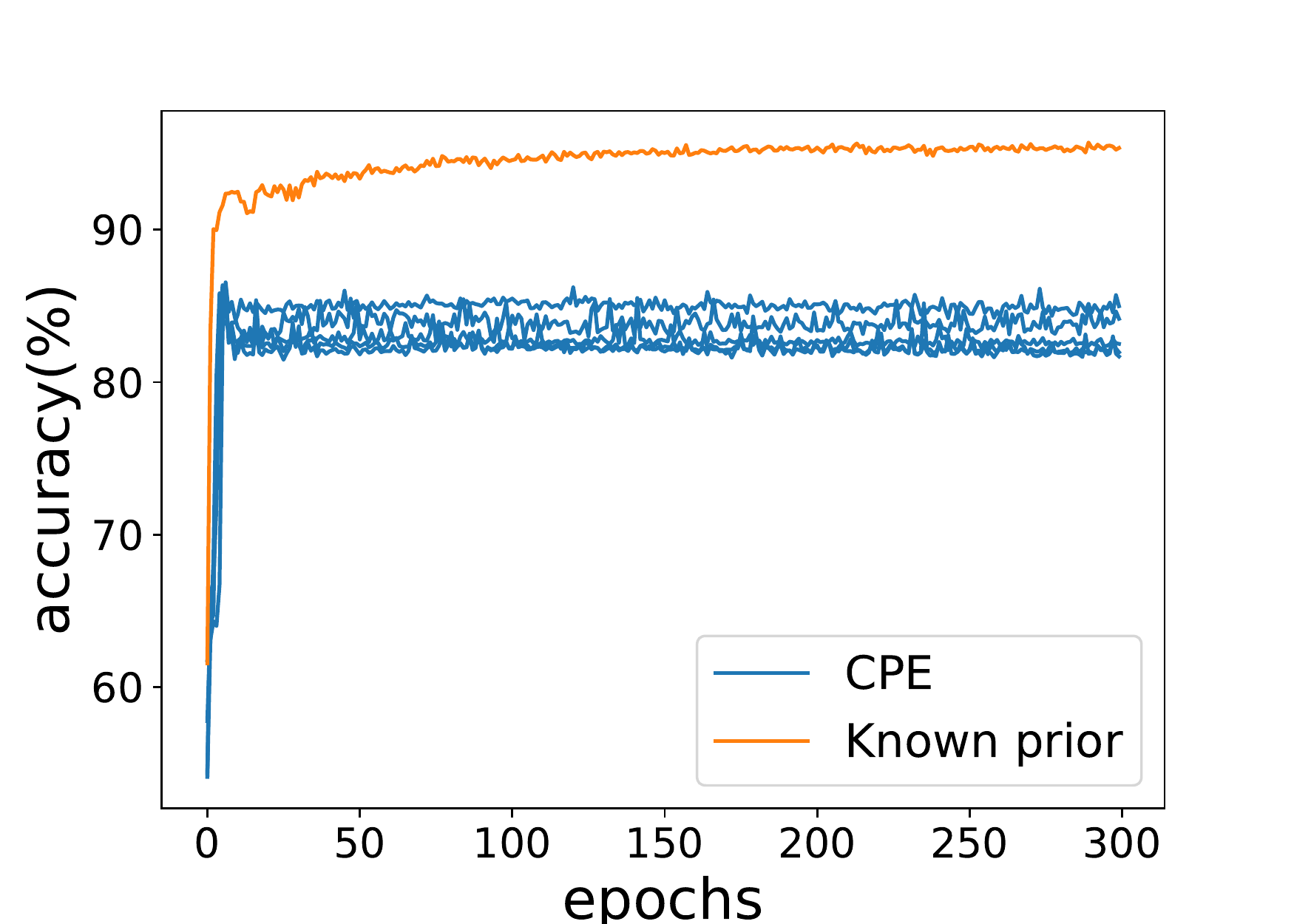}
        \caption{AWA2 (CPE)} 
        \label{fig.S3.3}
        \end{subfigure}%
    \begin{subfigure}{0.45\linewidth}
        \centering
        \includegraphics[width=1.12\linewidth]{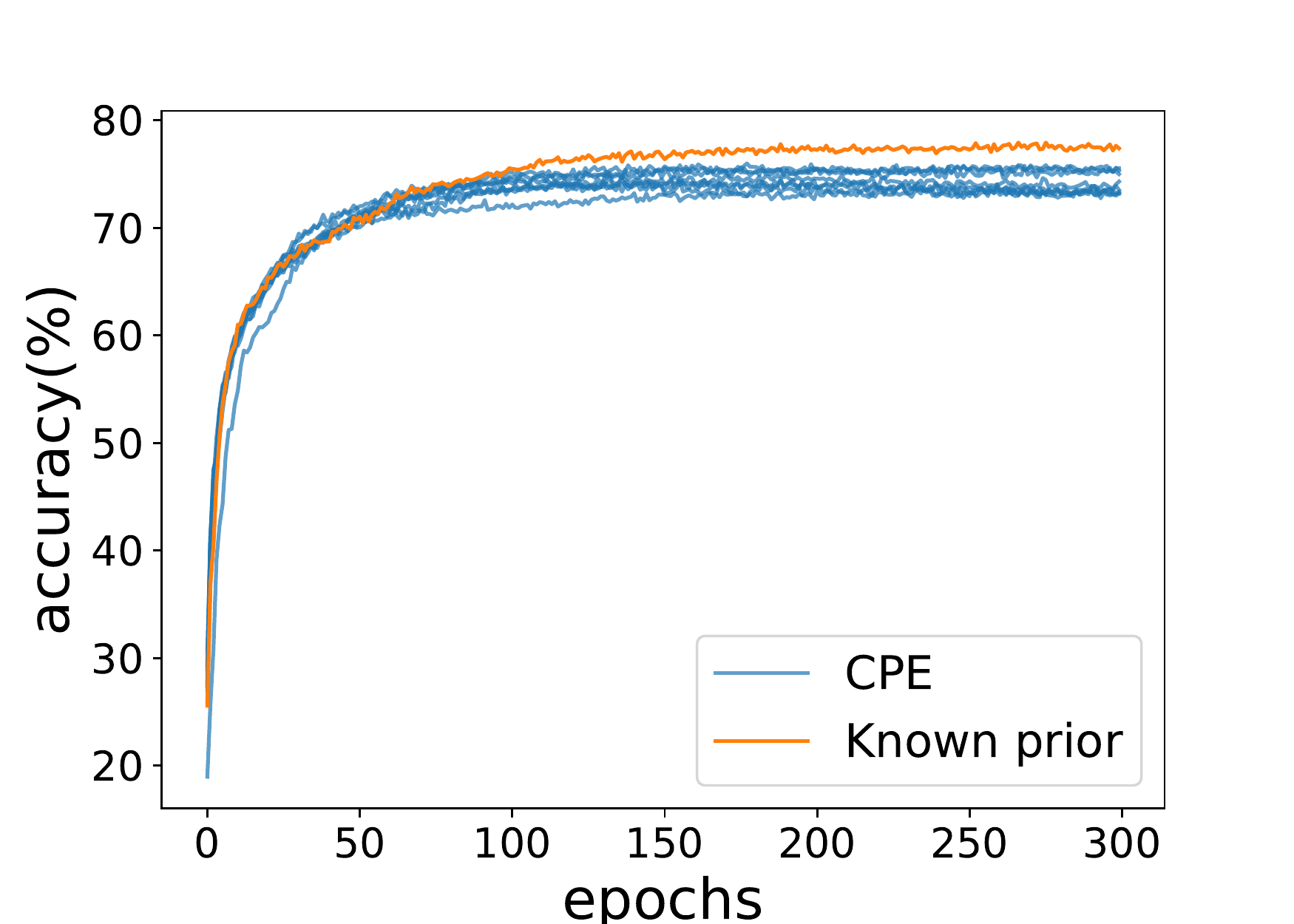}
        \caption{CUB (CPE)} 
        \label{fig.S3.3}
        \end{subfigure}%
        \caption{Training accuracy of Bi-VAEGAN using different random seeds when class prior is unknown.}
\end{figure}
Here we explain the  black box shift estimation  (BBSE) \cite{lipton2018detecting} approach for class prior estimation.
It attempts to solve the problem in label shift setting \cite{DBLP:conf/nips/GargWBL20} and we consider the TZSL problem in a discrete form i.e., $y \in Y^u = \{0,1,2,..., N_u - 1\}$.
We view our synthesized joint distribution $p^u_G(\hat{\bm v},y)$ as the source domain and the unknown joint distribution $p^u(\bm v,y)$ as the target domain.
Under the label shift assumption, i.e., $p_G^{u}(\hat{\bm v}|y) = p^u(\bm v|y)$, we can approximate the unseen prior via the normalized confusion matrix $\bm{C}_{\hat{y},y} := p^u_{\bm G}(\hat{y}|y)$ of synthesized features, where $\hat{y} = f(\hat{\bm v})$ is the predicted label using hypothesis $f$.
Following \cite{lipton2018detecting}, when the label shift condition is held and the confusion matrix is invertible, the following equation holds,
\begin{align}
\nonumber
     p^u(\hat{y}) =& \sum\limits_{y\in Y^u} p^u(\hat{y}|y)p^u(y) =\sum\limits_{y\in Y^u} p^u_{\bm G}(\hat{y}|y)p^u(y),\\
      =&\sum\limits_{y\in Y^u} \bm{C}_{\hat{y},y}{p^u(y)},
\end{align} 
thus $p^u(y)$ is computed as,
\begin{align}
    p^u(y) = &\sum\limits_{\hat{y}\in Y^u} \bm{C}^{-1}_{y,\hat{y}}{p^u(\hat{y})}.
\end{align}
To compute the confusion matrix $\bm{C}$ we synthesize two labeled unseen set $\langle \hat{V}^u_{\bm G},\hat{Y}^u_{\bm G} \rangle_1$ and $\langle \hat{V}^u_{\bm G},\hat{Y}^u_{\bm G} \rangle_2$.
We train the hypothesis on one labeled set and compute the confusion matrix on the other set. Note that as the training process goes, the confusion matrix tends to be an identity matrix and the BBSE estimation collapse to $p^u(y) \leftarrow p^u(\hat{y})$.

\begin{figure}[tp]
    \centering
    \begin{subfigure}{0.48\linewidth}
        \centering
        \includegraphics[width=1.12\linewidth]{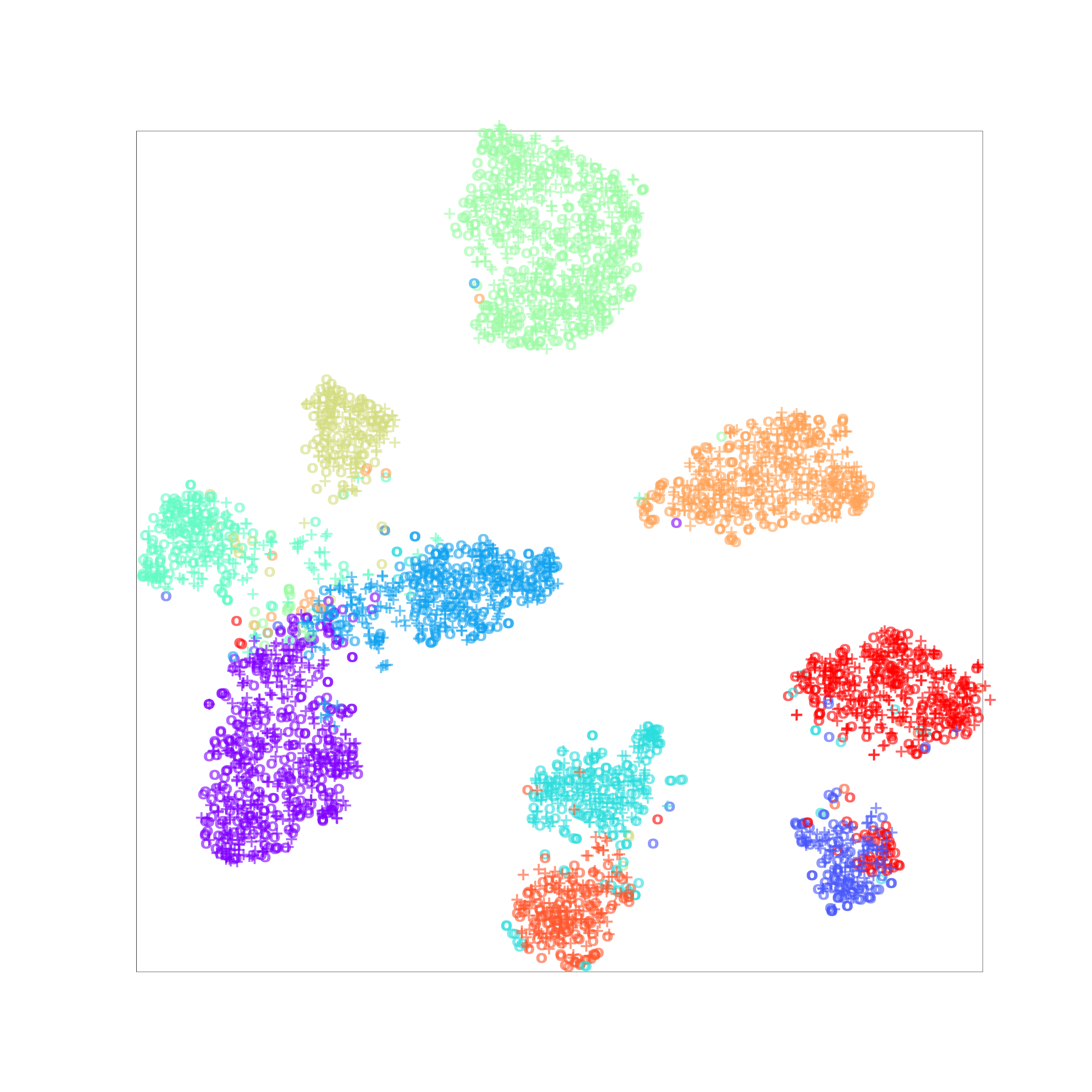}
        \caption{AWA1 + CPE (T1-91.5\%)} 
        \label{fig.S4.1}
    \end{subfigure}
    \begin{subfigure}{0.48\linewidth}
        \centering
        \includegraphics[width=1.12\linewidth]{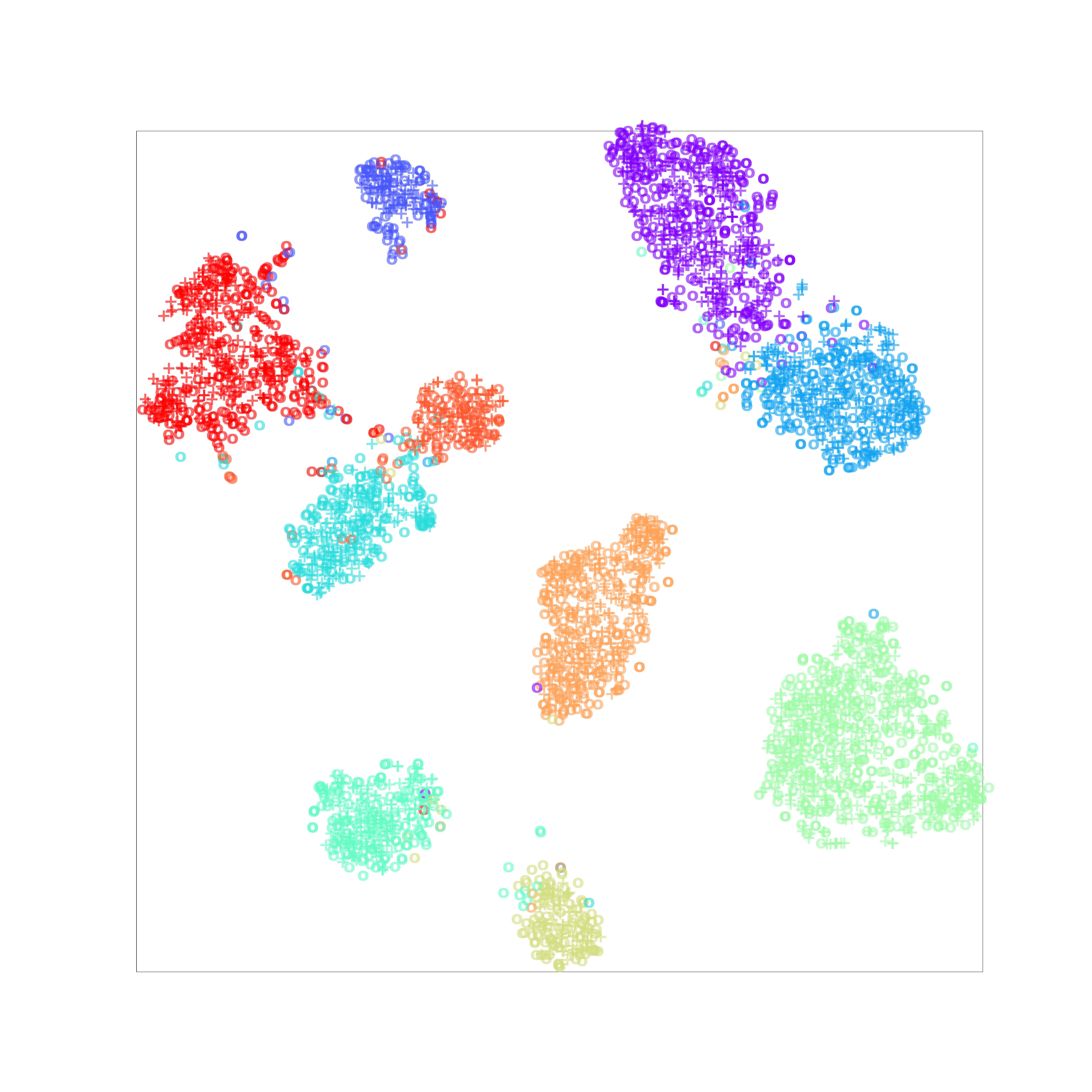}
        \caption{AWA1 (T1-93.9\%)} 

        \label{fig.S4.2}
    \end{subfigure}

    \begin{subfigure}{0.48\linewidth}
        \centering
        \includegraphics[width=1.12\linewidth]{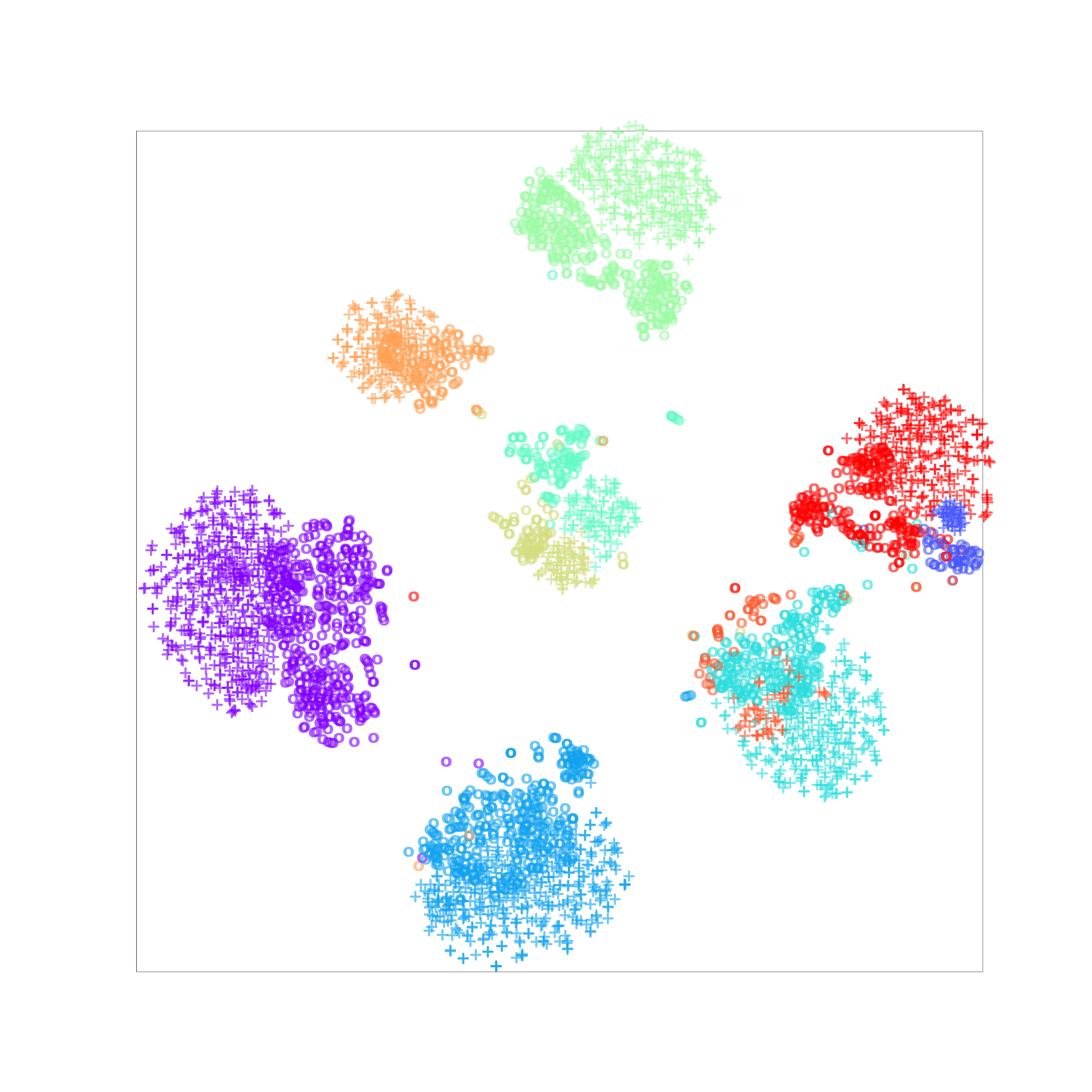}
        \caption{AWA2 + CPE (T1-85.6\%)} 
        \label{fig.S4.3}
    \end{subfigure}
    \begin{subfigure}{0.48\linewidth}
    \centering
    \includegraphics[width=1.12\linewidth]{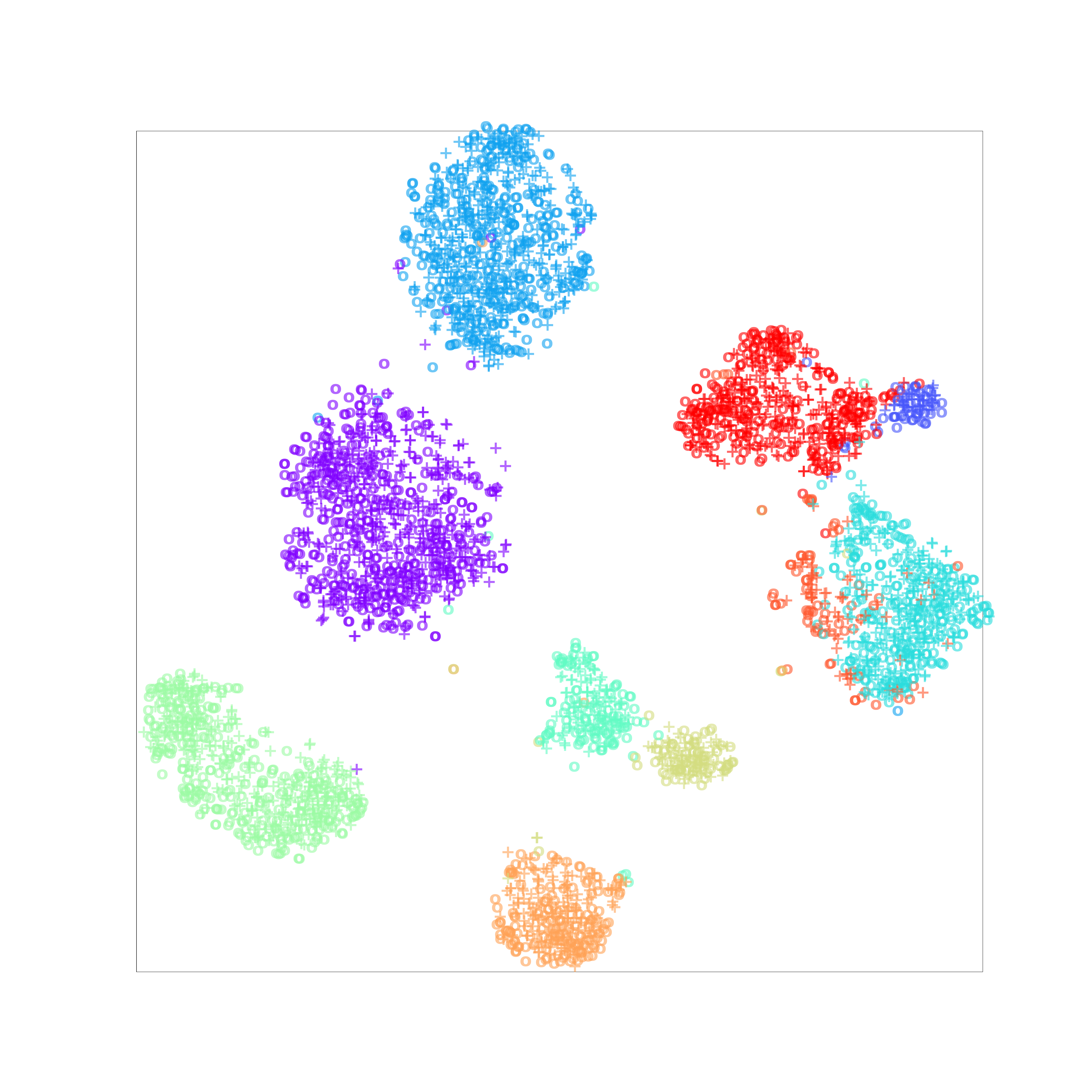}
    \caption{AWA2 (T1-95.8\%)} 
    \label{fig.S4.4}
    \end{subfigure}%
    \caption{Visualization of real/synthesized unseen visual feature using Bi-VAEGAN. Left column uses CPE strategy and class prior is unknown. The right column is trained with given real class prior. `$\circ$' means the real feature and `+' means the syntehsized feature.\label{fig.S4}}
\end{figure}

{\bf{Discussion.}} 
We display the BBSE and our CPE's training accuracy curves on AWA1 in Figure \ref{fig.S3.1} and Figure \ref{fig.S3.2}. It might be observed that BBSE is more vulnerable to seed selection and that it more readily results in a poor convergence. 
This observation can be explained as that the label shift assumption is too strong for prior estimation, so that the neural network classifier performs more unstably. 
The non-parametric K-means technique tends to provide a more moderate and reliable estimation since CPE avoids directly employing the black-box neural network classifier and utilizes it as an initialization approximation of the class center instead.

Figure \ref{fig.S4} shows the t-SNE visualizations using CPE when the class prior is unknown. For the more evenly balanced AWA1, our CPE provides a satisfactory alignment between the real and the syntehsized features, and there is only a minor accuracy gap with the known prior scenario (91.5\% vs. 93.9\%).
For the more unbalanced AWA2 dataset, the domain between the synthesized and real features shifts noticeably, and the performance disparity with the know-prior scenario increases to (85.6\% vs. 95.8\%). 
This supports the argument  of Corollary 3.1 that the class prior is crucial to the alignment of the conditional distribution for the TZSL.
However, it is still unclear how to proceed with a more accurate class prior estimation when the real prior is highly unbalanced. Different from the widely studied problems of  \textit{covariate shift} and \textit{label shift} in domain adaptation \cite{DBLP:conf/icml/0002CZG19, lipton2018detecting}, the unknown prior TZSL is less well-organized and is more similar to a cross-modal \textit{generalized label shift} problem. 

\end{document}